\documentclass[twoside,11pt]{article}

\usepackage{blindtext}

\usepackage[preprint]{jmlr2e}

\newcommand{\hangin}{\vspace{5pt}\goodbreak\hangindent=.50cm \noindent }

\usepackage{amsmath,amsfonts,amssymb,mathtools}
\usepackage{booktabs} % for professional tables
\usepackage[dvipsnames]{xcolor}
\usepackage{algorithm}
\usepackage{algpseudocode}
\usepackage{comment}
\hypersetup{hidelinks}

\newcommand{\etal}{\textit{et al}.}
\newcommand{\ie}{\textit{i}.\textit{e}., }

\def\a{\alpha}

\def\l{\lambda}

\firstpageno{1}

\usepackage{lastpage}
\jmlrheading{24}{2023}{1-\pageref{LastPage}}{3/23; Revised
6/23}{7/23}{23-0367}{Khurram Javed, Haseeb Shah, Richard S. Sutton, Martha White}
\ShortHeadings{Columnar-Constructive Networks}{Javed, Shah, Sutton, and White}

\begin{document}

\title{Scalable Real-Time Recurrent Learning Using \\ Columnar-Constructive Networks}

\author{\name Khurram Javed \email kjaved@ualberta.ca \\
        \name Haseeb Shah \email hshah1@ualberta.ca \\
        \name Richard S. Sutton \email rsutton@ualberta.ca \\
        \name Martha White \email whitem@ualberta.ca \\
       \addr Alberta Machine Intelligence Institute (Amii)\\Department of Computing Science, University of Alberta, \\ Edmonton, AB, Canada\\
    }

\editor{George Konidaris}

\maketitle

\begin{abstract}%
Constructing states from sequences of observations is an important component of reinforcement learning agents. One solution for state construction is to use recurrent neural networks. Back-propagation through time (BPTT), and real-time recurrent learning (RTRL) are two popular gradient-based methods for recurrent learning. BPTT requires complete trajectories of observations before it can compute the gradients and is unsuitable for online updates. RTRL can do online updates but scales poorly to large networks. In this paper, we propose two constraints that make RTRL scalable. We show that by either decomposing the network into independent modules or learning the network in stages, we can make RTRL scale linearly with the number of parameters. Unlike prior scalable gradient estimation algorithms, such as UORO and Truncated-BPTT, our algorithms do not add noise or bias to the gradient estimate. Instead, they trade off the functional capacity of the network for computationally efficient learning. We demonstrate the effectiveness of our approach over Truncated-BPTT on a prediction benchmark inspired by animal learning and by doing policy evaluation of pre-trained policies for Atari 2600 games. 
\end{abstract}


\section{Recurrent Networks for State Construction}
Learning by interacting with the world is a powerful framework for building systems that autonomously achieve goals in complex worlds. A key ingredient for building such systems is agent-state construction---learning a compact representation of the history of interactions that helps in predicting and controlling the future. One solution for state construction is to use differentiable recurrent neural networks (RNNs) learned to minimize prediction error~(Kapturowski~\etal,~2018; Vinyals~\etal,~2019).

State construction using neural networks requires structural credit assignment---identifying how to change network parameters to improve predictions. In RNNs, parameters influence predictions made many steps in the future, and credit assignment involves tracking the influence of the parameters on these future predictions. Two popular algorithms for gradient-based structural credit assignment are back-propagation through time (BPTT)~(Werbos, 1988; Robinson and Fallside, 1987)  and real-time recurrent learning (RTRL)~(Williams and Zipser,~1989). 

We define real-time state construction as the ability of the agent to learn the agent-state in real-time while interacting with the world. The agent does not postpone learning to the future by storing data, nor does it have access to specialized hardware that is taken away at the time of deployment. Instead, it has a fixed amount of computational resources throughout its lifetime, and there is no distinction between learning and deployment. 

Neither BPTT nor RTRL are suitable for real-time state construction. BPTT stores all past network activations and does sequential operations proportional to the length of the data stream for estimating the gradient. As a result, it neither scales well nor learns in real-time. RTRL does not require more per-step computation for longer sequences. However, it scales poorly to larger RNNs. Both BPTT and RTRL can be approximated for real-time learning with large networks.

A promising direction to scale gradient-based learning is to approximate the gradient. Elman~(1990) proposed to ignore the influence of parameters on future predictions. The resulting algorithm is computationally cheap but biased. Williams and Peng~(1990) proposed Truncated-BPTT (T-BPTT), an algorithm that tracks the influence of a parameter on predictions made up to k steps in the future, where k is a hyperparameter called the truncation length. T-BPTT works well on many benchmarks~(Mikolov~\etal, 2009, 2010; Sutskever, 2013 and Kapturowski~\etal, 2018), but cannot reliably learn associations beyond its truncation length (Mujika~\etal, 2018). Tallec~\etal~(2018) demonstrated T-BPTT can even diverge when observations have negative long-term effects on a target and positive short-term effects. 

Hochreiter~\&~Schmidhuber~(1997) used a diagonal approximation to RTRL (Diagonal-RTRL) that scales linearly with the number of parameters. Menick~\etal~(2021) generalized the diagonal approximation with their algorithm called SnAp-$k$. Diagonal-RTRL and SnAp-$1$ are not blind to all long-term dependencies, but introduce bias in the gradient estimate. They assume that changing a recurrent feature will not change the values of other features, an assumption that does not hold in densely connected recurrent networks. SnAp-$k$ for $k>1$ is less biased, but scales poorly. Tallec~\etal~(2017) proposed UORO, a computationally efficient algorithm for getting unbiased samples of gradients. However, the resulting samples are noisy and only effective for learning with small step-sizes. Menick~\etal~(2021) showed that UORO performs poorly even on simple benchmarks. 

Existing methods for scaling gradient-based recurrent learning approximate the gradient but do not make assumptions about the function class of the recurrent network. In this work, we propose a different strategy: we propose to limit the function class of the RNNs such that we can estimate the gradient efficiently and without bias.

\begin{figure*}[t]
	\centering
	\includegraphics[width=0.99\textwidth]{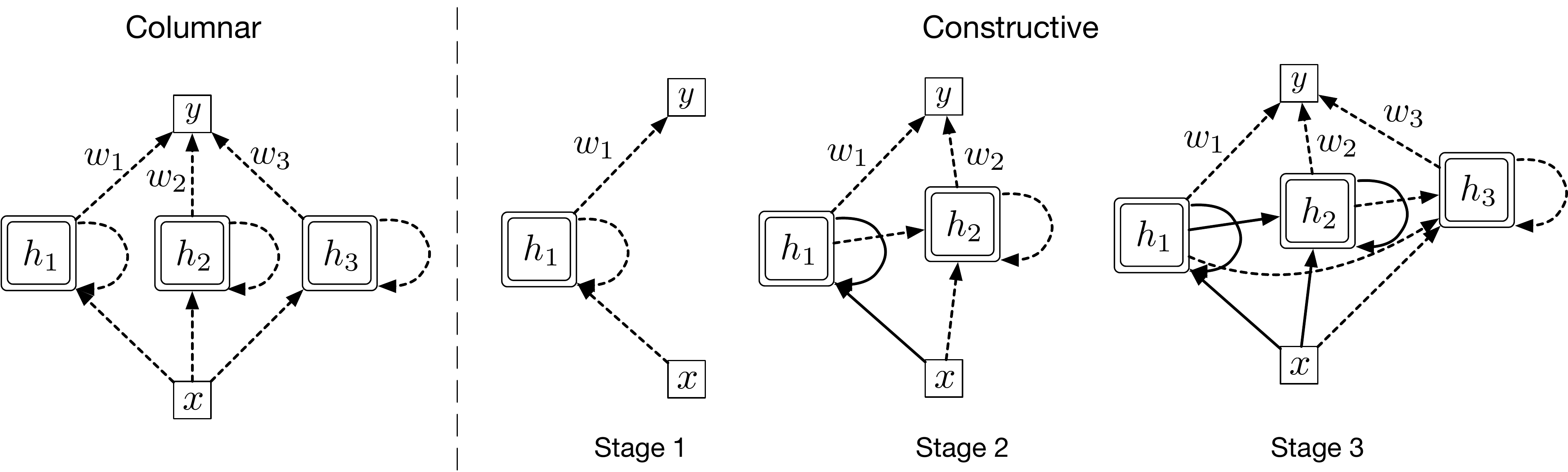}
	\caption{Two families of recurrent networks for which gradients can be efficiently computed without bias or noise. Recurrent networks with a columnar structure use $O(n)$ operations and memory per step for learning. However, they do not have hierarchical recurrent features---recurrent features composed of other recurrent features. Constructive networks introduce hierarchical recurrent features and learn them in stages to keep learning computationally efficient.}
	\label{two_approaches}
\end{figure*}
We first propose Columnar networks. They are composed of independent, potentially deep, columns. Each column has a scalar recurrent state, and is independent of other columns. As a result, the gradient of each recurrent feature is non-zero w.r.t parameters of exactly one column. The RTRL update for Columnar networks is computationally efficient: linear instead of quadratic in the number of parameters. Columnar networks lack hierarchical recurrent features---recurrent features composed of other recurrent features.

To introduce hierarchy in features, we propose Constructive networks. Constructive networks learn recurrent features in stages. In each stage, they freeze the existing recurrent features and learn a single new recurrent feature. They can learn hierarchical recurring features efficiently. However, they only update one feature at any given time, which can be slow.

Finally, to overcome the limitations of Columnar and Constructive networks, we propose Columnar-Constructive networks (CCNs). CCNs learn the network in stages, similar to Constructive networks. However, instead of learning one feature at a time, they learn multiple independent columns of features in a single stage, similar to Columnar networks. CCNs overcome the primary limitations of both Columnar and Constructive networks.

We compare CCNs to RNNs trained using T-BPTT and find that CCNs are more computationally efficient. As we increase the truncation length of T-BPTT, its performance improves; however, it also uses more computation and memory. We find that for the same per-step computation budget, CCNs perform better than T-BPTT, especially when relatively small agents have to interface with large and complex environments.

We evaluate our algorithms on two partially observable benchmarks to estimate values (prediction). First, we use an existing animal-learning benchmark~(Rafiee~\etal,~2022), which has low-dimensional inputs and a focus on the need for memory---the only way to make accurate predictions is to remember information from many steps in the past. Second, to test the algorithms in more complex image-based environments, we make a new benchmark based on ALE~(Arcade Learning Environment)~(Bellemare~\etal,~2013). We use policies of pre-trained Rainbow-DQN agents~(Hessel~\etal,~2018 and Fujita~\etal,~2021). Removing frame-stacking~(Mnih~\etal,~2015), and frame-skipping in ALE makes the environments partially observable. We down-scale the observations to make partial observability even more pronounced. Our prediction benchmark based on ALE is available \href{https://github.com/khurramjaved96/atari-prediction-benchmark}{here}. 

\section{Problem Formulation}
We formulate the goal of a learner as predicting the discounted sum of a cumulant from an online stream of experience. The agent observes $\textbf{x}_t\in \mathbb{R}^n$ at time $t$ and predicts a scalar $y_t$. The goal of the agent is to minimize the sum of squared error between the prediction and the discounted sum of future values of a cumulant c, where c is a fixed index of $\textbf{x}$, \ie the agent aims to minimize:
\begin{equation}
\begin{aligned}
\mathcal L(T-k, T) = \frac{1}{k} \sum_{t=T-k}^{T} (y_t - \sum_{j=t+1}^\infty \gamma^{j-t-1} c_j)^2
\label{goal}
\end{aligned}
\end{equation}

\begin{equation}
\begin{aligned}
\mathcal L(T) = \frac{1}{T} \sum_{t=0}^{T} (\hat{y}_t -  y_t)^2
\label{goal}
\end{aligned}
\end{equation}

where $k$, and $T$ control the horizon over which the prediction error is accumulated. Note that the error is measured w.r.t the predictions made over time and not using a final set of weights. 

Our problem formulation can capture various online temporal-prediction and supervised-learning problems. For example, setting the cumulant to the reward turns our problem into policy evaluation~(Sutton~\&~Barto,~2018). Setting $\gamma = 0$, the problem can represent online supervised recurrent learning benchmarks. The parameter $k$ allows us to smoothly move between the lifetime and final performance of the agent; by setting $k$ to a small value, we can measure the performance of the learner at the end of learning; similarly, by setting $k$ to $T$, we can measure the lifetime performance of the learner.

Traditionally, learning performance is evaluated on a held-out test set. While the train and test distinction is important in offline learning, when the learner has access to the complete data set, it is unnecessary when it sees the data online and is always evaluated on the next unseen data point \textit{before} using it for learning. 
% MARTHAC: Too soon. We haven't even said what a policy is
%Note that for all our experiments, the data generation policies $\pi$ are fixed and so do not need to be explicitly considered in the loss.

\subsection{Learning with Resource Constraints}
We focus on the under-parameterized setting where the environments are more complex than the learners. The learners have a fixed per-step compute and memory budget, that they can allocate however they choose. For instance, a learner can pick an expensive learning algorithm, such as RTRL, and satisfy the compute constraint by using a smaller recurrent network. Alternatively, it can choose a larger recurrent network, and learn the network using a computationally efficient learning algorithm, such as T-BPTT with a small truncation length. 

The focus on the under-parameterized setting, online learning, and real-time learning emphasizes the need of computationally efficient learning algorithms that can be applied continually. Moreover, since the real-world is significantly more complex compared to even the largest recurrent networks, the under-parameterized setting is arguably a better proxy for real-world problems. 

\subsection{Learning with Recurrent Architectures} 
In our temporal prediction setting, it is natural to assume that the learner will not fully observe the state of the environment. Instead, it might need information from its history of observations for making accurate predictions. Throughout this work, we assume that the learner attempts to summarize its history using recurrent neural networks (RNNs).  

The dynamics of an RNN can be written as 
\begin{equation}
\textbf{h}_t = f\left(\textbf{h}_{t-1}, \textbf{x}_t, \theta \right),
\end{equation}
where $\textbf{h}_t\in\mathbb{R}^d$ is the state of the network, $\textbf{x}_t$ is the observation, $\theta$ are the learnable parameters, and $f$ is the dynamics function of the network. 
The recurrent state $\textbf{h}_t$ is linearly weighted with weights $\textbf{w}_t\in\mathbb{R}^d$ to make a prediction $y_t$ as:
\begin{equation}
y_t = \sum_{k=0}^{d-1} h_{t,k} w_{t,k}
\end{equation}
where  $h_{t, k}$ and $w_{t, k}$ are the kth element of vectors $\textbf{h}_t$ and $\textbf{w}_t$, respectively. 

To update the parameters $\theta_t$ at time $t$, we need the gradient of the prediction with respect to $\theta$. Using the chain rule, we can write the gradient as
\begin{equation}
\begin{aligned}
\frac{\partial{y_t}}{\partial{\theta}} & = \frac{\partial{y_t}}{\partial{\textbf{h}_t}}\frac{\partial{\textbf{h}_t}}{\partial{\theta}}.   
\label{final_step}
\end{aligned}
\end{equation}
% where 
% \begin{equation}
%     \frac{\partial y_t}{\partial \theta_k} :=  \frac{\partial y_t}{ \partial \textbf{h}_{k}}\frac{\partial \textbf{h}_{k}}{\partial \theta} - \frac{\partial y_t}{\partial \textbf{h}_{k-1}}\frac{\partial \textbf{h}_{k-1}}{\partial \theta}
% \end{equation}
The key question is how to compute $\frac{\partial{\textbf{h}_t}}{\partial{\theta}}$. We can obtain a recursive formula for this expression, which is used by RTRL and by the algorithms we introduce in this work. To make it clear how we can use the multivariable chain rule, let us explicitly write 
$\textbf{h}_t(\theta) = f\left(\textbf{h}_{t-1}(\theta), \textbf{x}_t, \mathbf{g}_t(\theta) \right)$
where $\mathbf{g}_t(\theta) \doteq \theta$. Then the multivariable chain rule gives us
%We can expand the second term in equation~\ref{final_step} using the recursive relation: 
\begin{equation}
\frac{\partial{\textbf{h}_t}}{\partial{\theta}} =  \frac{\partial{\textbf{h}_t}}{\partial{\mathbf{g}_t}}\frac{\partial \mathbf{g}_t}{\partial \theta} + \frac{\partial{\textbf{h}_t}}{\partial{\textbf{h}_{t-1}}} \frac{\partial{\textbf{h}_{t-1}}}{\partial{\theta}},
\label{expansion_1}
\end{equation}
where the first term in the sum is the gradient of the state of the network under the assumption that $\textbf{h}_{t-1}$ is not a function of $\theta$, and the second term captures the indirect impact of $\theta$ on $\partial{\textbf{h}_t}$ due to its impact on $\partial{\textbf{h}_{t-1}}$.  

This recursive relationship is exploited by two algorithms: BPTT and RTRL.   
BPTT stores the network activations and inputs from prior steps and expands equation~\ref{final_step} as:
\begin{equation}
\begin{aligned}
\frac{\partial{y_t}}{\partial{\theta}}  &=  \frac{\partial{y_t}}{\partial{\textbf{h}_t}}\frac{\partial{\textbf{h}_t}}{\partial{\theta}} \\
\frac{\partial{y_t}}{\partial{\theta}}  &= \frac{\partial{y_t}}{\partial{\textbf{h}_t}}\frac{\partial{\textbf{h}_t}}{\partial{\mathbf{g}_t}}\frac{\partial \mathbf{g}_t}{\partial \theta} + \frac{\partial{y_t}}{\partial{\textbf{h}_t}}\frac{\partial{\textbf{h}_t}}{\partial{\textbf{h}_{t-1}}} \frac{\partial{\textbf{h}_{t-1}}}{\partial{\theta}}\\
&= \frac{\partial{y_t}}{\partial{\textbf{h}_t}}\frac{\partial{\textbf{h}_t}}{\partial{\mathbf{g}_t}}\frac{\partial \mathbf{g}_t}{\partial \theta} + \frac{\partial{y_t}}{\partial{\textbf{h}_t}}\frac{\partial{\textbf{h}_t}}{\partial{\textbf{h}_{t-1}}} \frac{\partial{\textbf{h}_{t-1}}}{\partial{\textbf{g}_{t-1}}}\frac{\partial \textbf{g}_{t-1}}{\partial \theta} + \frac{\partial{y_t}}{\partial{\textbf{h}_t}}\frac{\partial{\textbf{h}_t}}{\partial{\textbf{h}_{t-1}}} \frac{\partial{\textbf{h}_{t-1}}}{\partial{\textbf{h}_{t-2}}}\frac{\partial \textbf{h}_{t-2}}{\partial \theta} 
\end{aligned}
\label{expanded}
\end{equation}
 to compute the gradient. It unrolls the formula back in time, computing and accumulating gradient until the start of the recursion at $t = 0$. RTRL, on the other hand, updates the Jacobian $\frac{\partial \textbf{h}_k}{\partial \theta}$ using equation~\ref{expansion_1} at every step. 
To get the gradient w.r.t the prediction, it uses equation~\ref{final_step}. 

Both algorithms compute the same gradient, but make different compromises in terms of computation and memory. RTRL does not require storing past activations and inputs, as it can update the Jacobian using only the most recent input. However, computing the Jacobian using equation~\ref{expansion_1} requires  $O(|\textbf{h}|^2|\theta|)$ operations and $O(|\textbf{h}||\theta|)$ memory. The size of the parameters $|\theta|$ in a fully connected RNN is $|\textbf{h}|^2$. RTRL is therefore often said to have quartic complexity in the size of the hidden state. 
BPTT requires $O(|\theta|t)$ memory and compute, where $t$ is the length of the sequence. It avoids the bigger memory cost by computing the product $\frac{\partial{y_t}}{\partial{\textbf{h}_t}}\frac{\partial{\textbf{h}_t}}{\partial{\mathbf{g}_t}}\frac{\partial \mathbf{g}_t}{\partial \theta}$ directly, rather than separately computing the Jacobian and then multiplying by $\frac{\partial{y_t}}{\partial{\textbf{h}_t}}$. For sequences shorter than $|\textbf{h}|^2$, BPTT is cheaper than RTRL for fully connected RNNs. 

\section{Columnar-Constructive Networks} 

\begin{figure}[t]
	\centering
	\includegraphics[width=0.85\textwidth]{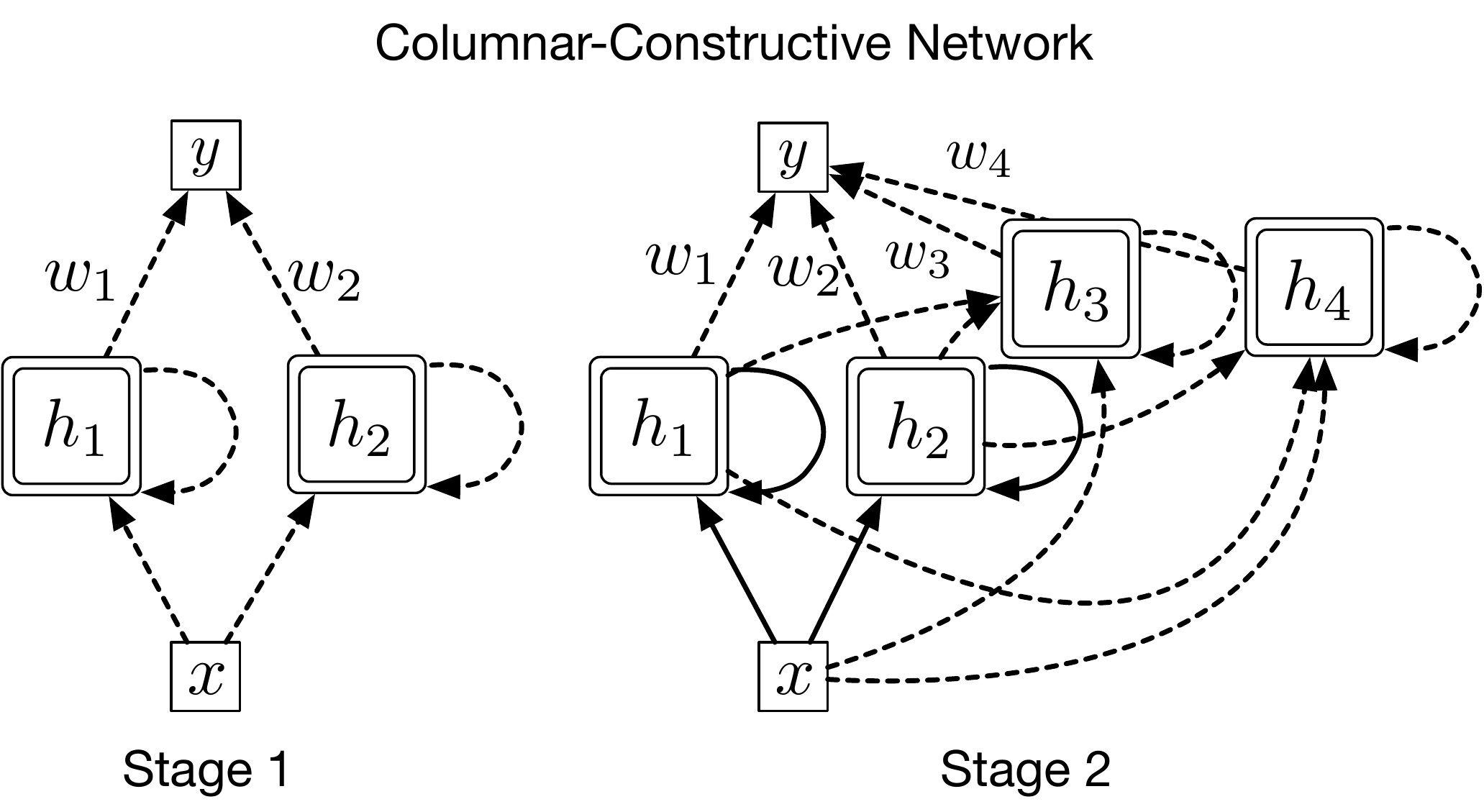}
	\caption{Columnar-Constructive networks (CCNs) combine the ideas from Columnar and Constructive networks. In each stage, the they learn multiple features that are independent of each other, just like Columnar networks. Across stages, they learn hierarchical features, similar to the Constructive networks.}
	\label{hybrid_picture}
\end{figure}

In this section we develop a new approach for recurrent learning, called Columnar-Constructive networks (CCNs). CCNs leverage two key ideas: First, RTRL is computationally efficient for modular recurrent networks where each module has a scalar hidden state; we call these networks Columnar networks. Second, RTRL is computationally efficient if the recurrent units are learned in stages, as opposed to simultaneously. We call the incremental learning approach Constructive networks. Figure~\ref{two_approaches} visualizes the central ideas behind Columnar and Constructive networks

Both Columnar and Constructive networks, on their own, show promising results but have limitations. Columnar networks cannot learn hierarchical features, and Constructive networks cannot learn multiple features in parallel. We show that their weaknesses can be overcome by combining the two ideas to create a third system that we call Columnar-Constructive networks (CCNs). 

\subsection{Columnar Networks}
Columnar networks organize the recurrent network such that each scalar recurrent feature is independent of other recurrent features. Let $h_{t, k}$ be the kth index of the state vector $\textbf{h}_t$. Then, in columnar networks,
\begin{equation}
h_{t, k} = f_k(h_{t-1,k}, \textbf{x}_t, \theta_{t, k}).
\end{equation}
Each $f_k$ outputs a scalar recurrent feature and is called a column.\footnote{This terminology comes from the connection to structure observed in brains~(Mountcastle~1957).} $\theta_{t, k}$ is the set of parameters of the $kth$ column. For any $i\ne j$, the set $\theta_{t, i}$ and $\theta_{t, j}$ are disjoint. A columnar network consists of $d$ columns. The output of all columns are concatenated to get the $d$-dimensional hidden-state vector $\textbf{h}_t$. Figure~\ref{two_approaches}~(left) shows a graphical representation of a Columnar network. Note that changing $h_1$ has no influence on the value of $h_2$ or $h_3$.

Because recurrent features in a columnar network are independent of each other, we can apply RTRL to each of them individually. 
To better understand why, let us rederive our recursive formula for the gradient. For $\theta_k$, the parameters for the $k$th column, we have
\begin{equation*}
\frac{\partial{y_t}}{\partial{\theta_k}} = \frac{\partial{y_t}}{\partial{\textbf{h}_t}}\frac{\partial{\textbf{h}_t}}{\partial{\theta_k}} 
= \sum_{j=1}^d \frac{\partial{y_t}}{\partial{{h}_{t,j}}}\frac{\partial{{h}_{t,j}}}{\partial{\theta_k}} 
= \frac{\partial{y_t}}{\partial{{h}_{t,k}}}\frac{\partial{{h}_{t,k}}}{\partial{\theta_k}}. 
\end{equation*}
All except one term in the summation above are zero because $\theta_k$ does not influence $h_{t,j}$ when $j\ne k$. Therefore, we only have to compute $\frac{\partial{{h}_{t,k}}}{\partial{\theta_k}}$ with RTRL. Like before, we can write this recursively using
${h}_{t,k}(\theta_k) = f\left({h}_{t-1,k}(\theta_k), \textbf{x}_t, \mathbf{g}_t(\theta_k) \right)$
where $\mathbf{g}_t(\theta_k) \doteq \theta_k$, giving 
\begin{equation}
\frac{\partial{{h}_{t,k}}}{\partial{\theta_k}} = 
\frac{\partial{{h}_{t,k}}}{\partial{\mathbf{g}_t}}\frac{\partial \mathbf{g}_t}{\partial \theta_k} + \frac{\partial{{h}_{t,k}}}{\partial{{h}_{t-1,k}}} \frac{\partial{{h}_{t-1,k}}}{\partial{\theta_k}}.
\label{expansion}
\end{equation}
Computing and storing this Jacobian only costs $O(|\theta_{t, k}|)$ memory and compute for each column because $|h_{t,i}| = 1$ for a single column. The cost for all the columns is 
\begin{equation}
    O(|\theta_{t, 1}|) + O(|\theta_{t, 2}|) + \cdots + O(|\theta_{t, n}|) = O(|\theta_t|).
\end{equation}
Therefore, RTRL for Columnar Networks scales linearly in the size of the parameters. 
In this work, we implement each column as a single LSTM cell with a hidden size of one. We provide the explicit gradients in Appendix \ref{update_equations}. 

\subsection{Constructive Networks}
In Constructive networks, we learn the recurrent network one feature at a time. Features learned later can take as input all features learned before them; the opposite is not allowed---feature learned earlier cannot take as input features that would be learned later. We elucidate the multi-stage learning process in a small Constructive network in Figure~\ref{two_approaches}~(right). Dotted lines represent parameters that are being updated at every step, whereas solid lines represent parameters that are fixed. 

In the first stage, the learner learns the incoming weights of $h_1$, which is connected to the input features $x$, but not to $h_2$ or $h_3$. Note that we are omitting the time index for brevity, and $h_1$ is the same as $h_{t,1}$. Once the incoming and the recurrent weights of $h_1$ are learned, the learner freezes them and goes to stage 2. In stage 2, it learns the incoming weights of $h_2$. $h_2$ can use both $x$ and $h_1$ as the inputs. The outgoing weight of $h_1$---$w_1$---is not fixed and continues to be updated. Similarly, in the 3rd stage, both $h_1$ and $h_2$ are frozen and fed to $h_3$ as input. In each stage, the newly introduced feature can be connected to all prior features.

In this staged learning approach, the learner never learns more than one feature at a time. As a result, the effective size of the hidden state of the learning system is just one, and RTRL can be applied cheaply. In fact, since only a small subset of the network is being learned at any given time, Constructive networks use even less per-step computation than Columnar networks. They introduce one additional hyperparameter---steps-per-stage---that controls the number of steps after which the learner moves from one stage to the next. 

Constructive networks are similar to prior work on recurrent cascade correlation networks~(Fahlman, 1990). The main differences are that (1) cascade correlation networks learn new recurrent units by maximizing correlation with the error whereas Constructive networks use the gradient w.r.t the prediction error and (2) cascade correlation networks learn on a batch of data, whereas Constructive networks learns from an online stream of data. The two differences are arguably minor. Rather, the bigger novelty is to combine Constructive networks with Columnar networks, as discussed in the next section.

\subsection{Columnar-Constructive Networks}

Columnar-Constructive networks (CCNs), as the name suggests, are a combination of Columnar and Constructive networks. In CCNs, we keep the multi-stage approach of the Constructive networks; however, instead of learning a single feature in every stage, the learner learns multiple independent features.

A two-stage CCN is shown in Figure~\ref{hybrid_picture}. In stage one, the learner learns the incoming weights of $h_1$ and $h_2$. Since $h_1$ and $h_2$ are independent of each other, they are equivalent to a Columnar network with two features, and can be learned efficiently together. In the second stage, the learner freezes the incoming and recurrent weights of $h_1$ and $h_2$, and learns the incoming weights of $h_3$ and $h_4$; the new features take both $h_1$ and $h_2$ as inputs. Once again, $h_3$ and $h_4$ are independent of each other and can be learned efficiently in parallel. 

CCNs inherit the hyperparameters from Columnar and Constructive networks. Additionally, they have one new hyperparameter---features-per-stage---that controls the number of recurrent features learned in each stage

% Hybrid networks allow the learner to learn multiple recurrent features simultaneously, similar to columnar networks, while also learning hierarchical features in different stages, similar to constructive networks. 

\subsection{Feature Normalization}
A key to making our system work is online feature normalization. Unlike dense recurrent networks, features in our constructive and CCN networks can have varying number of incoming weights. This discrepancy can change the scale of each feature, making it hard to learn using a uniform step-size. To address the varying scales, we propose online feature normalization. Our feature normalization is similar to an online version of batch normalization~(Ioffe~and~Szegedy~2015). Prior work has shown feature normalization to be helpful for recurrent networks in the batch setting~(Cooijmans~\etal~2017).

To normalize a feature, we maintain an online running estimate of its mean and variance. We then use the running estimates to normalize the feature to have zero mean and unit variance. Additionally, if the variance of a feature goes below a threshold, we set it to a small number $\epsilon$---a hyperparameter---to prevent the normalized feature from getting too large. Capping the maximum value of the feature is important to prevent unstable behavior. 
Given the unnormalized feature $h_j$, the normalized feature $\hat{h}_j$ is computed as:
\begin{align}
    \hat{h}_{t, j} &= \frac{h_{t, j} - \mu_{t,j}}{\text{max}(\epsilon, \sigma_{t,j})} \\
 \text{where }   \mu_{t,j} &= \mu_{t-1,j}  \beta + (1-\beta) h_{t,j} \nonumber\\
    \sigma^2_{t,j} &= \sigma^2_{t-1, j} \beta + (1 - \beta) (\mu_{t, j} - h_{t,j})(\mu_{t-1, j} - h_{t,j}) \nonumber.
\end{align}
We set $\beta = 0.99999$ for all our experiments. $\mu_{0,i}$ and $\sigma^2_{0,i}$ are initialized to be 0 and 1 respectively. $\epsilon$ is tuned; the values used in this work are shown in Table~\ref{sweep_ranges} in Appendix \ref{app_hypers}. 
\section{Experiments on an Animal Learning Benchmark}
We start by evaluating the methods on a recently proposed benchmark inspired by animal learning (Rafiee~\etal, 2022). 
The trace patterning task (Rafiee~\etal, 2022) is an online prediction task that requires the learner to identify associations between vectors---conditional stimuli (CS)---and scalars---unconditional stimuli (US)---seen after a time delay. The goal is to predict the discounted sum of the US. Correct predictions require the ability to discriminate between patterns that lead to US from those that do not. The time delay between the CS and US necessitates retaining information from the past for making accurate predictions. 

In our instantiation of trace patterning, the delay between the CS and US is uniformly randomly sampled to be between 24 and 36 steps after every CS, and is called the inter-stimulus interval (ISI). The delay between the US and next CS is uniformly randomly sampled to be between 80 and 120 steps after every US, and is called the inter-trial interval (ITI). The CS consists of 6 features. When CS is present, three of the six features in the CS vector are one. Since $6 \choose 3$  is twenty, the CS vector can represent twenty different patterns. Ten randomly chosen patterns are followed by US=1 after ISI $\sim \mathcal U_{[24, 36]}$ steps, whereas the remaining ten do not activate the US. Additionally, the observation has five random features that are not predictive of the US. The learner has to ignore the random features to make accurate predictions.
\begin{figure}[ht]
	\centering
	\includegraphics[width=0.90\textwidth]{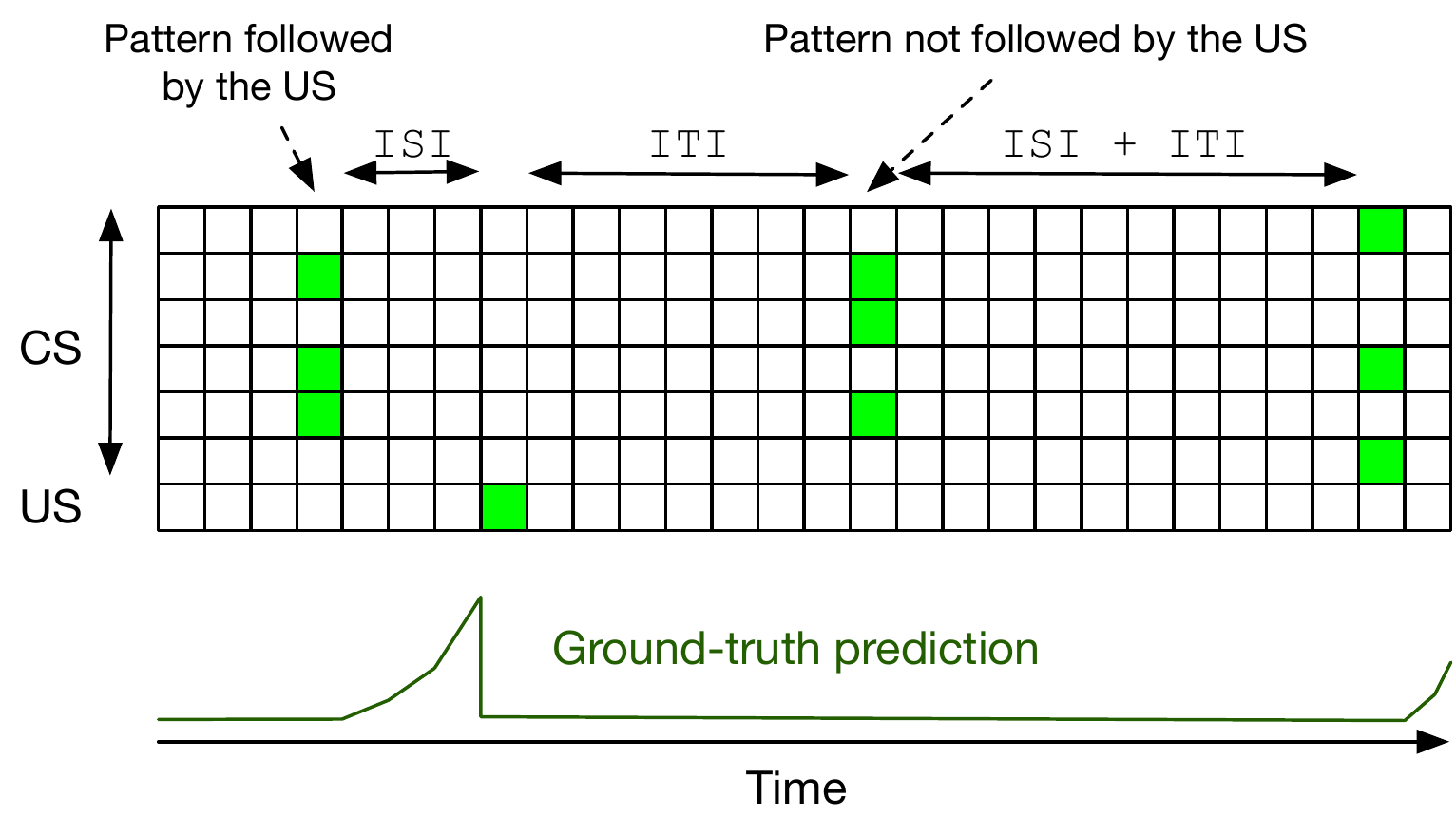}
	\caption{Visualization of the stream of experience for the trace patterning task. At each step, the learner sees a vector of length seven. The first six values are the CS, and the last is the US. CS is either a vector of zeros, or three of the six values are one. The CS can represent 20 different patterns. Ten of these patterns activate the US after ISI number of steps, whereas others do not. The learner has to predict the discounted sum of the value of US in the future. The bottom part of the figure shows the ground-truth prediction for the task.}
	\label{experience_image}
\end{figure}

A visual representation of experience from the trace patterning benchmark without random features with an ISI of 3 and ITI of 7 is shown in Figure~\ref{experience_image}. The vertical dimension shows the observations, and the horizontal represents time. At the fourth time-step, three of the six features are one. After three more steps, the US becomes active. Then no features are active for ITI number of steps. After ITI steps, the CS again becomes active. The second pattern of the CS is not followed by US. At the bottom of Figure~\ref{experience_image}, we show the ground truth return that the learner has to predict to minimize the prediction error. 

\subsection{Experimental Setup}
We compare CCNs to T-BPTT, Columnar networks and Constructive networks. All networks use the LSTM cell architecture~(Hochreiter~\&~Schmidhuber,~1997) for recurrence. For T-BPTT, we use a fully connected LSTM network. T-BPTT introduces another hyperparameter---the truncation length k. To keep the per-step computation constant, learners using a larger truncation lengths have fewer features. 

We use TD($\lambda$)~(Sutton,~1984, 1988; Tesauro, 1995) for learning. The full algorithm is in Appendix \ref{algo_pseudo}. We set the per-step compute budget to $\approx$ 4,000 floating point operations and treat multiplication, addition, division, and subtraction as one operation each.   We use $\lambda = 0.99$, and $\gamma = 0.90$ and report the learning curves for 10 million steps. At each point in the curve, we plot the error over the previous 100,000 data-points---we plot $\mathcal L(t-100,000, t)$ as a function of t.

For each method, we individually tune the step-size, $\epsilon$, steps-per-stage, features-per-stage, and the truncation length; we report the results for the best performing configuration. Details of hyperparameter tuning are in Appendix \ref{app_hypers}. The Columnar networks, Constructive networks, and CCNs have 10, 5, and 16 features respectively. The number of features in Constructive networks is dictated by the rate at which features are added. Because we only learn for 10 million steps, and use 1 million steps-per-stage, Constructive networks end up using significantly less compute than the allocated compute budget. T-BPTT uses a truncation length of 15, and has four features.
\begin{figure}[t]
	\centering
	\includegraphics[width=0.95\textwidth]{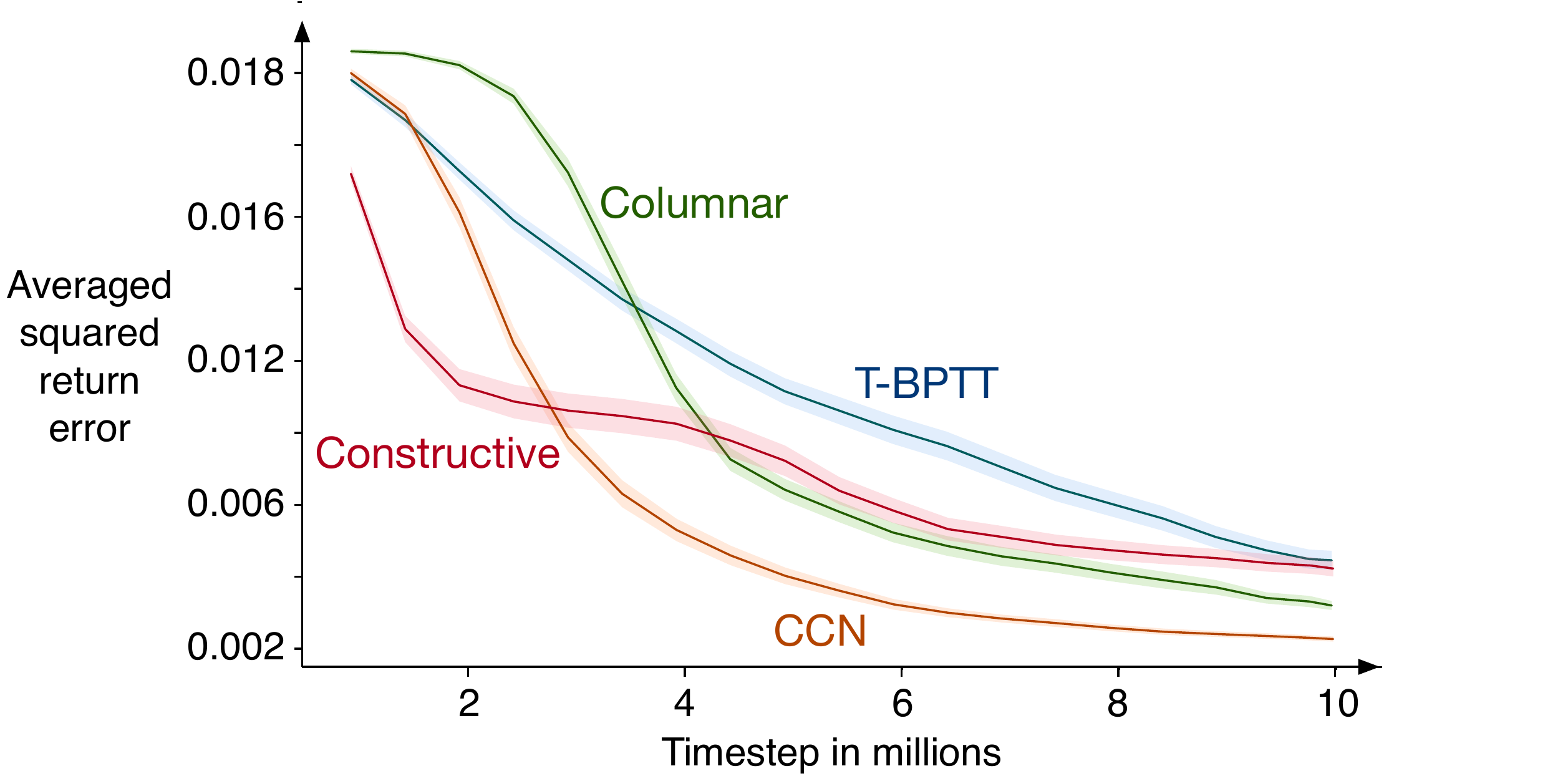}
	\caption{Performance of our algorithms, and the best performing T-BPTT on the trace patterning task. All methods can learn to make accurate predictions. Both Columnar and Constructive networks learn well, exceeding and matching the performance of best T-BPTT, respectively. CCNs combine the strengths of both and performs the best. All plots are averaged over 100 seeds, and the shaded areas are +- standard error.}
	\label{animal_results_main}
\end{figure}

\begin{figure}[t]
	\centering
	\includegraphics[width=0.95\textwidth]{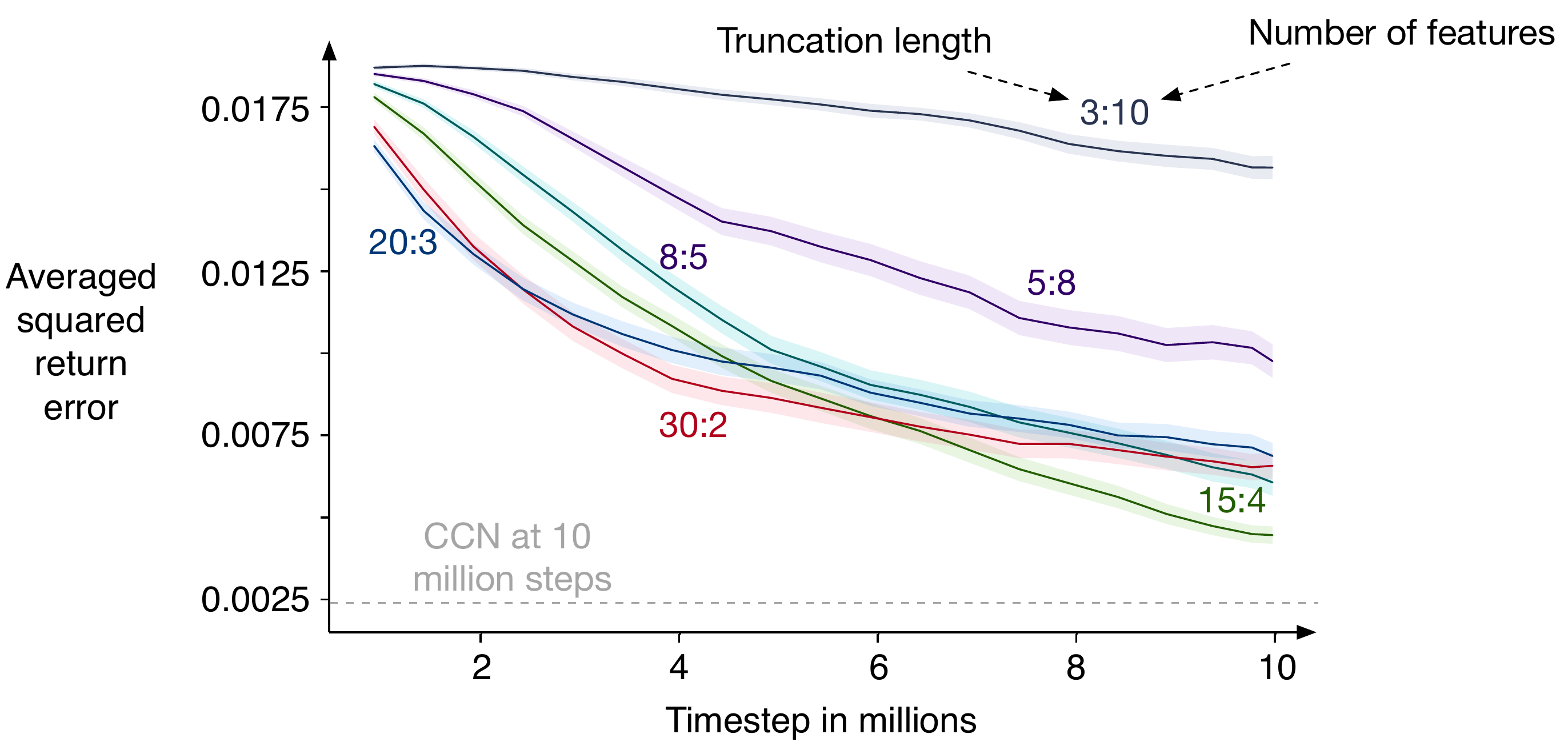}
	\caption{Different versions of T-BPTT on the trace patterning task. Each curve is denoted by two numbers: a:b. The first number indicates the truncation length of T-BPTT, and the second number indicates the number of features in the learner. For example, 30:2 means an LSTM with two features trained with a truncation length of 30. All versions use roughly the same amount of computation. We see that different values of truncation result in very different performances. Large networks trained with small truncation lengths---3:10 and 5:8---perform the worst showing the impact of the bias introduced by truncation. Smaller networks with longer truncation lengths---15:4, 30:2, and 20:3---perform better. All lines are averaged over 100 random seeds. The gray dotted line shows the performance of CCNs after learning for 10 million steps.}
		\label{tbptt_graph}
\end{figure}

\subsection{Results}

We start by looking at the learning curves for all four methods in Figure~\ref{animal_results_main}.
All three approaches learn to reduce the prediction error over time. Among our algorithms, Constructive networks perform the worst, demonstrating that learning one feature at a time is limiting. Both CCNs and Columnar networks reliably converge to a good solution and outperform the best T-BPTT. CCNs perform the best, demonstrating the usefulness of hierarchical features. 

We further investigate the sensitivity of T-BPTT to the value of truncation length. We first consider the impact of reallocating resources, allowing T-BPTT to have bigger networks with shorter truncation lengths and vice-versa. We see from Figure~\ref{tbptt_graph} that when the truncation length is much smaller than the longest dependency in the learning problem---36---the performance drops significantly. T-BPTT performs the best when it selects a smaller network (four features) and longer truncation ($k=15$).

\begin{figure}[t]
	\centering
	\includegraphics[width=0.95\textwidth]{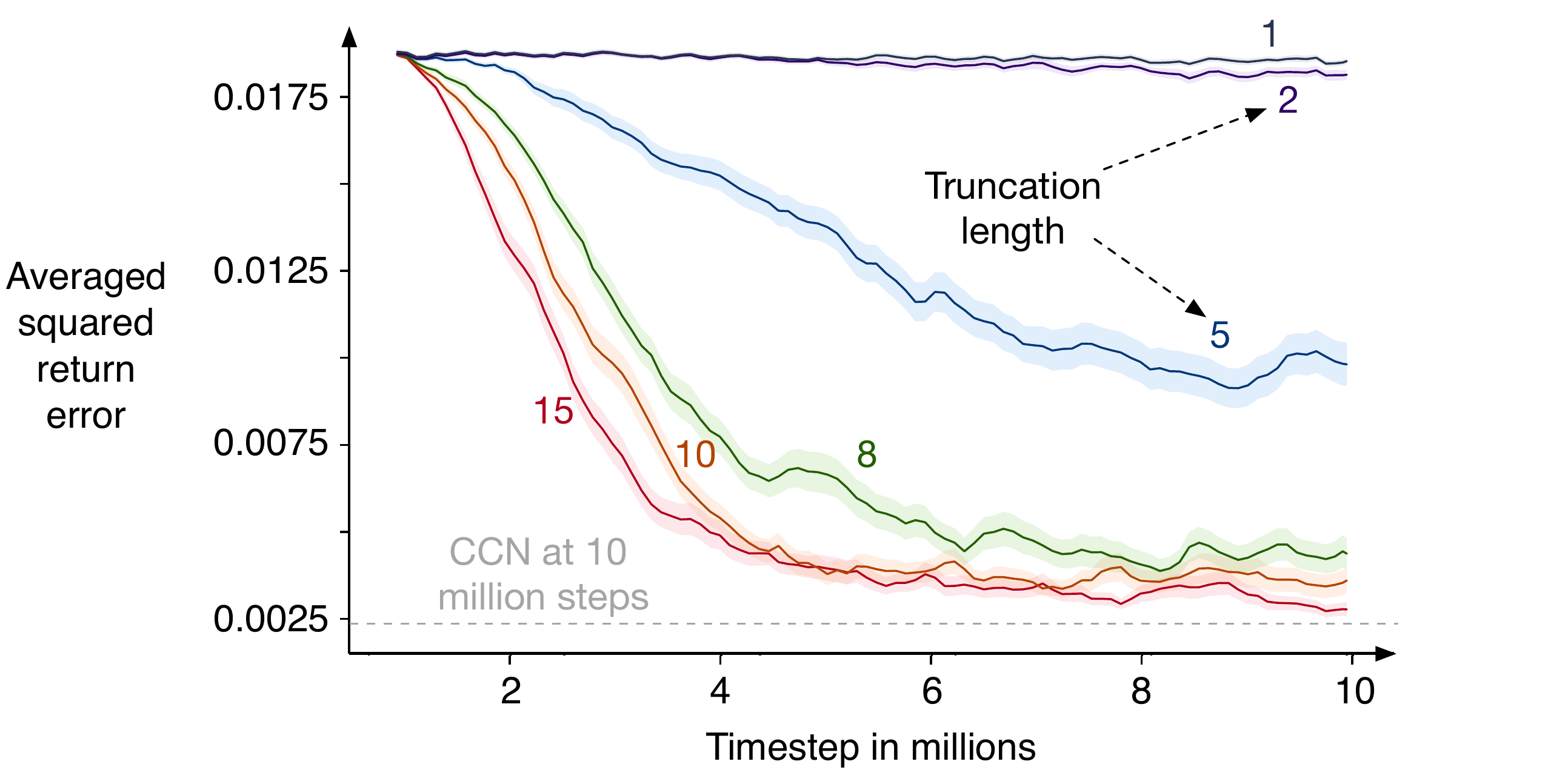}
	\caption{LSTMs with 10 hidden units trained using truncation lengths of 1, 2, 5, 8, 10, and 15. For each truncation length, we independently tune the step-size parameter. As the truncation length increases, the performance improves at the expense of more computation. The sensitivity of performance to truncation length highlights the impact of bias introduced by truncation. All lines are averaged over 100 random seeds and the shaded regions correspond to +- standard error. The gray dotted line shows the performance of CCNs after learning for 10 million steps.}
		\label{truncation_windoow}
\end{figure}

We conduct another experiment where we allow T-BPTT to use more computation than the allocated budget.
We fix the number of features to 10 and use different truncation lengths. We report the results in Figure~\ref{truncation_windoow}. Networks with the largest truncation length---red line---perform almost as well as CCNs. However, they use around seven times more per-step computation than CCNs.

\section{Experiments in the Arcade Learning Environment}
To evaluate our algorithms on more complex problems, we introduce a benchmark based on the Arcade Learning Environment~(Bellemare~\etal,~2013). Since our goal is to study state construction in the prediction setting, we use policy evaluation for pre-trained Atari policies, as opposed to reward maximization, as our benchmark.

We use the model zoo of Chainer-RL~(Fujita~\etal,~2021) to get pre-trained Rainbow-DQN~(Hessel~\etal,~2018) agents. For each game, we use the pre-trained agents to collect 50 million interactions. The policies use action-repeat (Machado \etal, 2018) and frame-skipping. Additionally, at the beginning of each episode, we take $i \sim \mathcal U_{[1, 30]}$ no-op actions to make the trajectories stochastic. We clip the rewards to be in the range $(-1, +1)$.

\begin{figure}[t]
	\centering
	\includegraphics[width=0.95\textwidth]{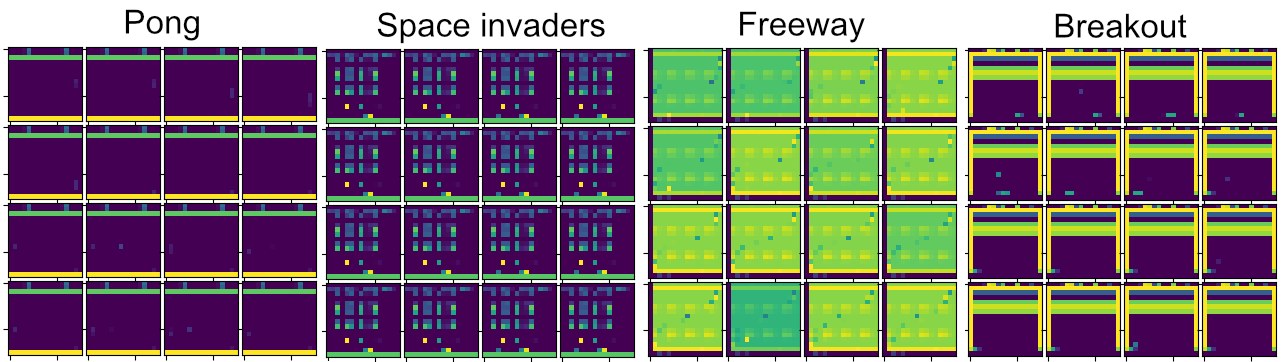}
	\caption{Environments down-scaled to 16 x 16. A single frame does not have sufficient information for making accurate predictions. For instance, in Pong, the ball is often not visible in a single frame. However, by looking at the sequence of frames, we can estimate the position and the direction of the ball. This partial observability due to down-scaling makes 16 x 16 Atari an interesting benchmark for studying state construction.}
		\label{envs_vis}
\end{figure}

Typically, Atari agents employ frame-stacking and frame-skipping to reduce partial observability. Since our goal is to study how well algorithms construct agent-states autonomously, we do not employ frame-stacking and frame-skipping. Additionally, we downscale the frames to $16 \times 16$---256 features---to make the partial observability even more pronounced. We visualize the downscaled observations of four games in Figure \ref{envs_vis} and see that single frames do not have sufficient information for making accurate predictions. Finally, we append the action and the previous reward to the game frame.

The final observation vector given to our learners has 275 features---256 values for the down-scaled frame, 18 for the one-hot encoded action, and one for the reward. 

\subsection{Experimental Setup}
We compare our methods to T-BPTT. All methods use LSTMs as the recurrent architecture and TD($\lambda$) for learning.

Each method has a per-step compute budget of $\approx$ 50k operations. We set  $\gamma$ to 0.98 and $\lambda$ to 0.99. The remaining parameters---$\epsilon$, steps-per-stage, truncation length, and step-size---are tuned independently for each method. The details of the hyperparameter tuning are in Table~\ref{app_hypers}. We pick hyperparameters that give the best results averaged over all the environments.

For each environment, we learn for 30 million steps and measure the average return error for the final 500k steps \ie $\mathcal L$(29,500,000, 30,000,000). Since different games have vastly different scales of returns, we normalize the errors before plotting them. For each game, we divide the error of all methods by the error achieved by T-BPTT in that game. As a result, the normalized error for the T-BPTT baseline is one in all games, whereas the performance of other methods is relative to that achieved by T-BPTT. For instance, a relative error of 0.5 for a method means that its error was half of T-BPTT.

\begin{figure*}
	\centering
	\includegraphics[width=0.99\textwidth]{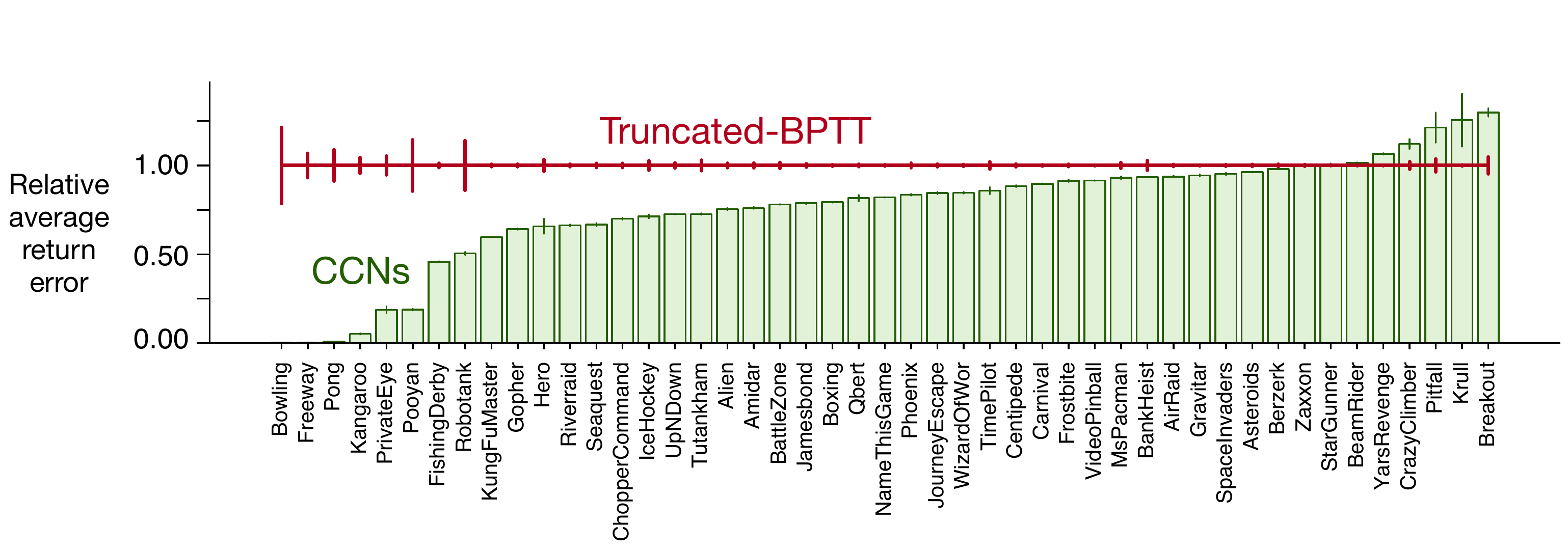}
	\caption{The performance of CCNs compared to the best T-BPTT on the Atari prediction benchmark. In most games, CCNs achieve lower prediction error than T-BPTT. In many games, CCNs reduce the prediction error by many folds. There are no games in which T-BPTT outperform CCNs by a large margin. All errors are averaged over 20 random seeds, and the error margins are +- standard error.}
	\label{atari_all_plot}
\end{figure*}

\begin{figure}[t]
	\centering
	\includegraphics[width=0.95\textwidth]{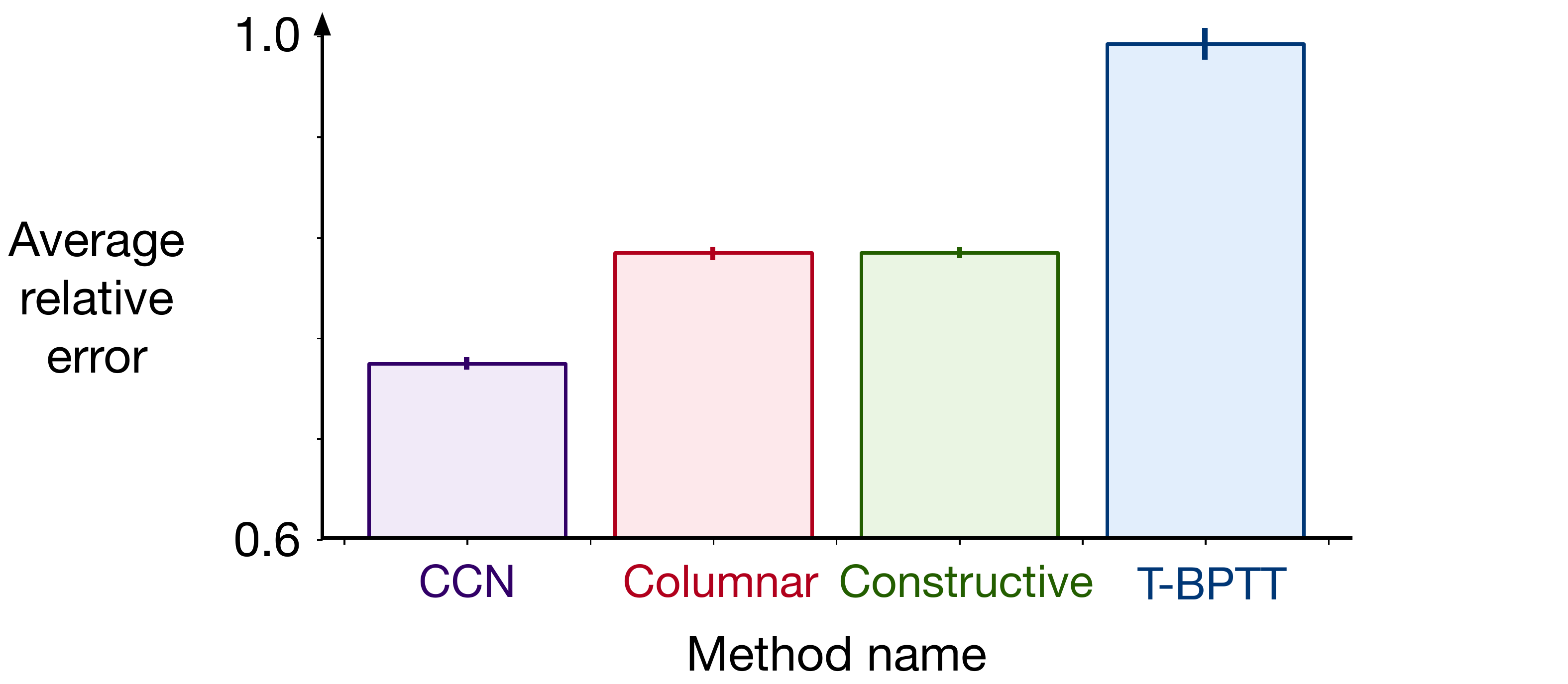}
	\caption{The averaged relative error achieved by the three proposed methods on the Atari prediction benchmark. Both Constructive and Columnar networks perform better than T-BPTT. CCNs perform the best, demonstrating they effectively combine the strengths of Columnar and Constructive networks. All results are averaged over 20 seeds and the error bars represent +- standard error.}
		\label{atari_more_details_1}
\end{figure}

\subsection{Results}

 We report the normalized error across all environments for CCNs and T-BPTT in Figure~\ref{atari_all_plot}. CCNs perform better than T-BPTT in most environments. In many environments, they achieve over 5x lower error, whereas even in the worst case of Breakout, the error is only 20\% more than T-BPTT.
 
We also look at errors achieved by Constructive and Columnar networks and report them in Figure~\ref{atari_more_details_1}. For brevity, we only report the average normalized error over all environments. All three of our methods improve over T-BPTT. CCNs perform the best, demonstrating that combining Columnar and Constructive networks is useful.

\begin{figure}[t]
	\centering
	\includegraphics[width=0.99\textwidth]{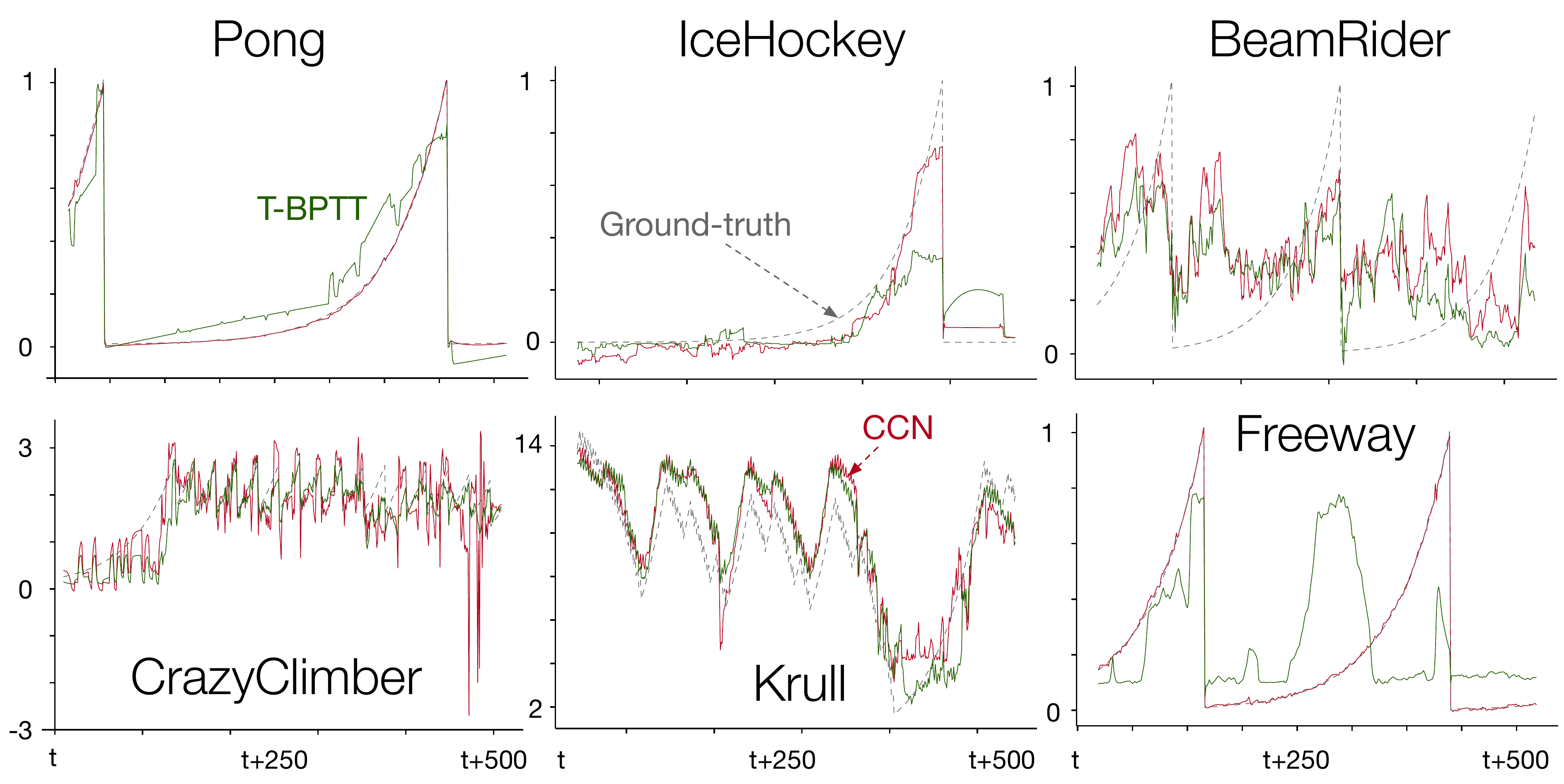}
	\caption{Predictions made by T-BPTT and CCNs on six Atari environments after learning for 30 million steps. The red lines are predictions made by CCN, the green lines are those made by T-BPTT, and the dotted grey lines are the ground truth returns. CCNs make qualitatively better predictions on most environments. The difference is most pronounced in Pong, and Freeway, in which CCNs make near-perfect predictions. T-BPTT achieves slightly lower error on CrazyClimber and Krull. Both algorithms struggle to make accurate predictions on BeamRider.}
		\label{preds}
\end{figure}

We visualize the predictions made by CCNs and T-BPTT at the end of learning in Figure~\ref{preds} on six environments. We pick three environments that favor CCN and three that favor T-BPTT. Both methods can learn to make accurate predictions. Predictions made by CCNs are closer, on average, to the ground truth returns than the predictions made by T-BPTT. The difference is most pronounced in Pong, and Freeway, where CCNs make near-perfect predictions. In BeamRider, both methods are unable to learn accurate predictions. One reason could be that the downsampled 16 x 16 frame does not have sufficient information for predicting the returns. Nonetheless, the Atari prediction benchmark provides significant evidence that CCNs outperform T-BPTT.

\subsection{Sensitivity of T-BPTT to Truncation Length}
To give a complete picture of the performance of T-BPTT,  we investigate the impact of the number of features and truncation length on its performance in two experiments.

\begin{figure}[t]
	\centering
	\includegraphics[width=0.85\textwidth]{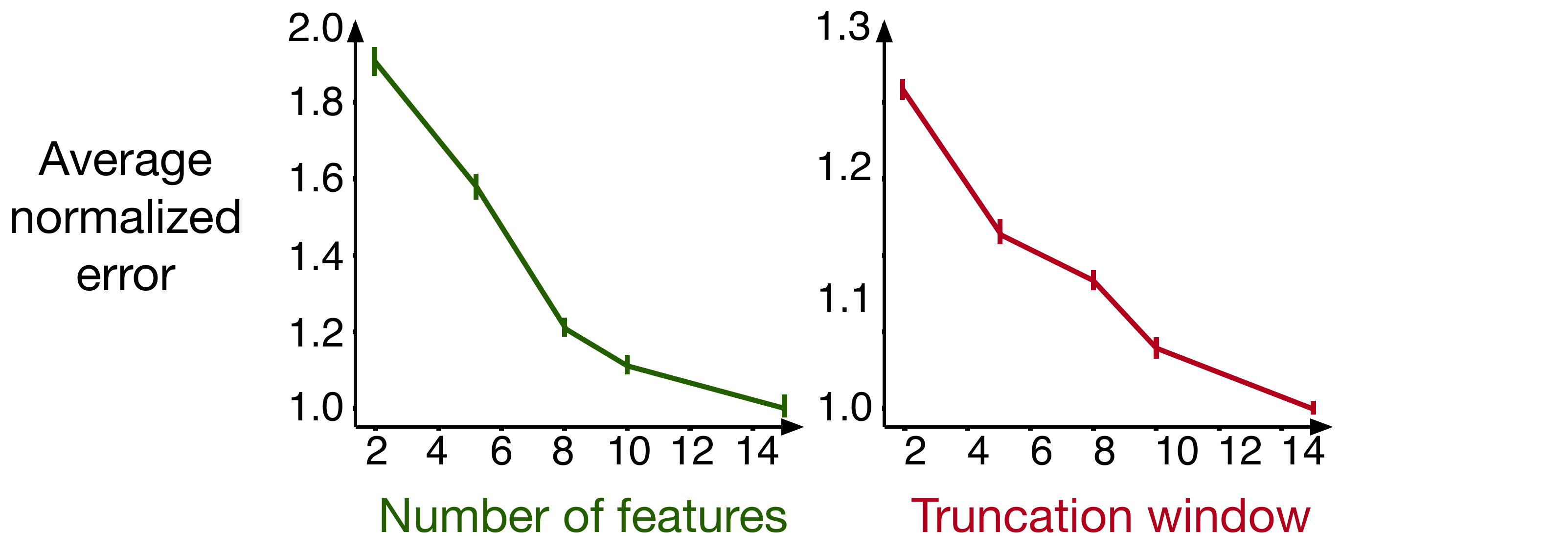}
	\caption{Impact of capacity and truncation length on the performance of T-BPTT on the Atari prediction benchmark. To generate the graph on the left, we fix the truncation length to 8, and vary the number of features. As the network gets larger, the performance improves. The error of an LSTM using two features is twice as much as an LSTM using 15 features. To generate the graph on right, we fix the number of features to 8, and vary the truncation length of T-BPTT. We see that bias introduced by smaller truncation length hurts performance. The errors in both plots are normalized such that the average error is one when number of features/truncation length is 15. }
		\label{atari_more_details_2}
\end{figure}

In the first experiment, we fix the truncation length to 8 and vary the number of features from 2 to 15. In the second experiment, we fix the number of features to 8 and vary the truncation length from 2 to 15. We report both results in Figure~\ref{atari_more_details_2}. We see that increasing both the number of features and the truncation length improves the performance of T-BPTT.

\section{Conclusions and Future Directions}

In this paper, we showed that by either restricting connections between recurrent neurons---Columnar networks---or learning a recurrent network in stages---Constructive networks---we can make RTRL computationally efficient. Moreover, unlike T-BPTT, our algorithms do not depend on sequential operations and can be parallelized. In the under-parameterized setting, our methods outperform T-BPTT. Moreover, they can learn networks with billions of parameters using roughly the same amount of resources needed for the deployment of similarly sized models. Because the learning algorithms are computationally efficient, there would be no need to disable learning at the time of deployment, and our systems could learn continually. 

One major limitation of our approach is that in both Constructive Networks and CCNs, most of the features are frozen as time goes by.  There are two routes to address this limitation. First, we can change the frozen features very slowly. The gradient should remain mostly accurate since these slowly changing features are effectively frozen from the timescale of the fast-changing features. Second, we can combine our approach with online weight and feature pruning. Instead of only adding features to grow the size of the network, we can instead continually replace the least useful features with new features, and learn them, as proposed by Dohare~\etal~(2023).  

Another open question in this work is to investigate if the restrictions introduced by CCNs make RNNs less general.
A theoretical understanding of the subclass of functions learnable by CCNs would help identify when, or even if, these networks might be insufficient for a given problem. Prior work (Giles~\etal,~1995 and Kremer, 1995) has analyzed the limitations of Recurrent Cascade-Correlation (RCC) networks, which are similar to Constructive networks. Kremer (1995) showed that RCC cannot learn certain Finite State Automata with linear threshold and sigmoid activations. It is not yet clear if CCNs suffer from similar problems; the argument used by Kremer~(1995) might not extend to the complex LSTM architecture used in our networks. Alternatively, the ability to learn multiple features in parallel might allow CCNs to not have the same limitations as RCC networks.

\acks{We are grateful to the Alberta Machine Intelligence Institute (Amii), CIFAR, and DeepMind for funding this research and to The Digital Research Alliance of Canada for providing computational resources. We are also thankful to the anonymous reviewers for improving the paper with their useful feedback.}


%\newpage

\appendix
\section{Hyperparameters and Ablations}
We provide details of the hyperparameters, and some ablatations in the following sections.
\subsection{Hyperparameter Settings}\label{app_hypers}
We tune the solution-specific hyperparameters for all the methods independently. For each configurations, we use five random seeds and look at the performance over all five seeds to pick the best hyperparameters. We then run the best hyperparameter configuration for 100 seeds for reporting the trace patterning results and 20 seeds for reporting the Atari results. List of all the hyperparameter, and their values are given in Table~\ref{sweep_ranges}.

\begin{table}[ht]
	\begin{tabular}{l l l l  }
		\toprule
		 Symbol &  Hyperparameter & Environment & Hyperparameter values\\ 
		 \midrule
		$\a$  & Step-size & All  & $ 1^{-2}, 3^{-3},1^{-3},  $ \\ & & & $3^{-4}, \mathbf{1^{-4}}, 3^{-5}$ \\
		 $\beta_1$:$\beta_2$:$\epsilon$ & Adam parameters  & All  & 0:0.9999:$1e^{-8}$ \\
		$\gamma$ & Discount factor & Trace & 0.90 \\
			$\gamma$ & Discount factor & Atari & 0.98 \\
		$\l$ & Eligibility trace decay rate & Both & 0.99              \\
		$k:d$    & Truncation:Hidden features (T-BPTT) & Trace & 2:13,
              3:10,
              5:8,
              8:5, \\ & & & 
              10:5,
              \textbf{15:4},
              20:3,
              30:2  \\
              $k:d$    & Truncation:Hidden features (T-BPTT) & Atari & 15:2, 8:5, \textbf{5:7}, \\ & & &  4:10, 2:25 \\
              & Hidden features (Columnar) & Trace &  5 \\
              & Hidden features (Columnar) & Atari &  6 \\
		& Features-per-stage (CCN) & Trace &  4 \\
			& Features-per-stage (CCN) & Atari &  5 \\
		 & Steps-per-stage (CCN)    &  Trace & 2.5 million \\
		  & Steps-per-stage (CCN)    &  Atari & 10 million \\
		& Steps-per-stage (Constructive) & Trace & 1 million \\
  & Steps-per-stage (Constructive) & Atari & 3 million \\
		& Total steps & Trace & 10 million \\ 
  & Total steps & Atari & 30 million \\ 
		& Seeds for parameter sweep & Both & $\{0, 1, 2, 3, 4\}$ \\
		& Seeds for best parameter configuration & Trace & $\{0, 1, \cdots, 99\}$ \\
		& Seeds for best parameter configuration & Atari & $\{0, 1, \cdots, 19\}$ \\
		$\epsilon$ & Min division term (CCNs and Constructive) & Both & $\{ 0.01, \textbf{0.001}\}$ \\
		\bottomrule
	\end{tabular}
 	\caption{Hyperparameter sweeps used for comparing algorithms}
	\label{sweep_ranges}
\end{table}

\subsection{TD Learning for Temporal Predictions}\label{algo_pseudo}
We use TD($\lambda$) for learning predictions as described in Algorithm~\ref{lambda_algo}. Every observation is processed once, and the target from the previous step is cached for computing the TD-error. Our implementation is similar to the one used in TD-Gammon~(Tesauro 1995). 
\begin{algorithm}
\label{lambda_algo}
\caption{TD($\lambda$) for online prediction}\label{alg:cap}
\begin{algorithmic}
\Require A differentiable learner $v_\theta$
\Require Step-size parameters $\alpha$
\Require Discount factor $\gamma$
\State Initialize \textbf{x} as first observation
\State Initialize eligibility  $\textbf{z}$ to \textbf{0}
\State $y = v_{\theta}(\textbf{x})$
\While{true}
\State Observe next $\textbf{x}^{'}$ and the cumulant $c$
\State $y^{'} = v_{\theta}(\textbf{x}^{'})$
\State $\delta = c + \gamma y^{'} - y$
\State $\textbf{z} = \lambda \gamma \textbf{z} + \nabla v_{\theta}(\textbf{x})$
\State $\theta  = \theta + \alpha \delta \textbf{z}$ 
\State $y = y^{'}$
\EndWhile
\end{algorithmic}
\end{algorithm}
\subsection{Impact of Normalization on The Performance of CCNs} 
\begin{figure}[t]
	\centering
	\includegraphics[width=0.95\textwidth]{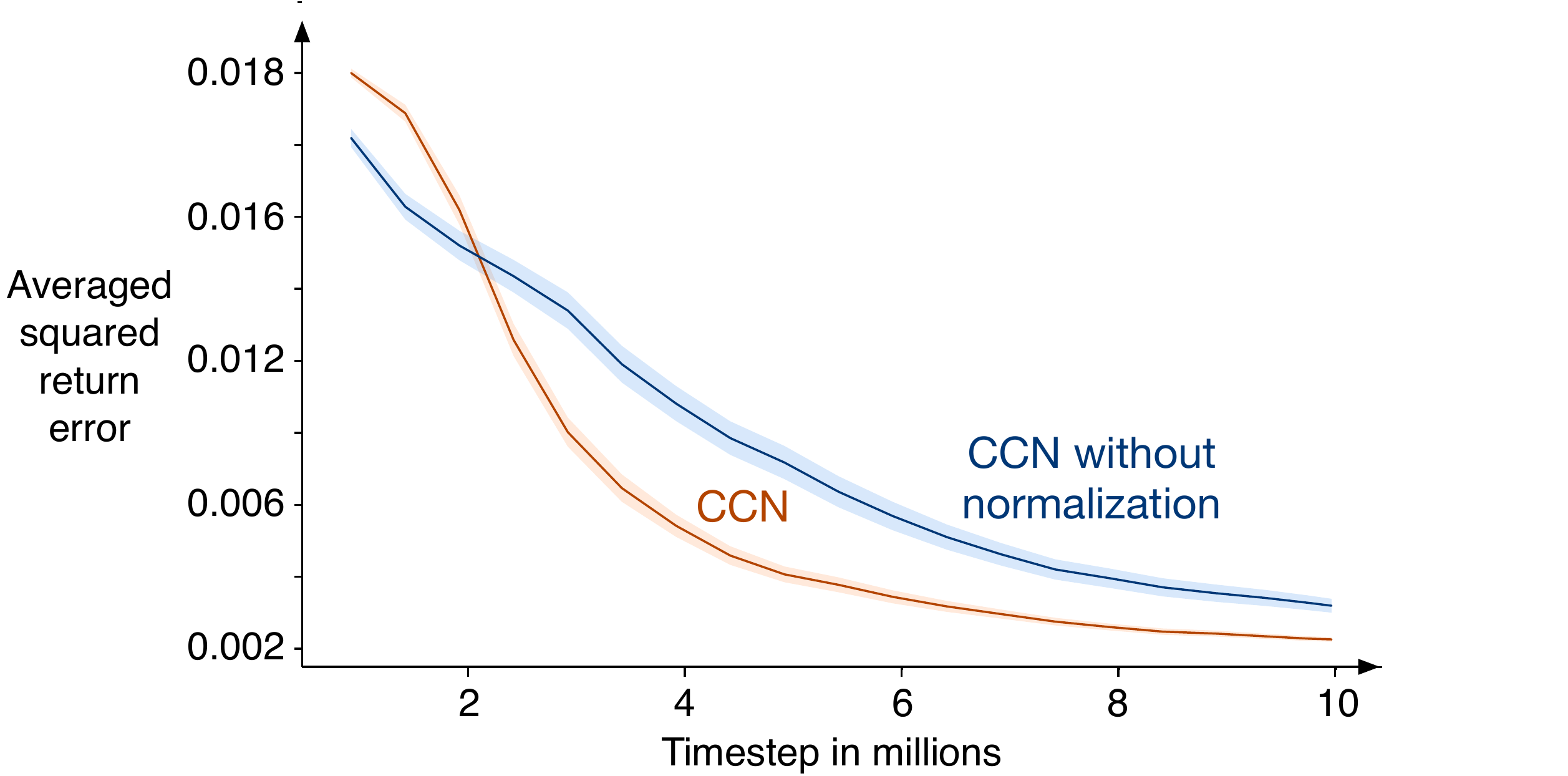}
	\caption{Impact of feature normalization on performance. The performance of CCNs is significantly worse when feature normalization is not used.}
		\label{norm_impact}
\end{figure}

We run ablations without the online feature normalization and find that feature normalization is indeed an essential component to make our networks work in a stable way. We compare the best performing CCNs with CCNs without normalization in Figure~\ref{norm_impact} and see that normalization improves performance. The normalized CCNs have both lower average error and variance. 

\begin{figure*}
	\centering
	\includegraphics[width=0.99\textwidth]{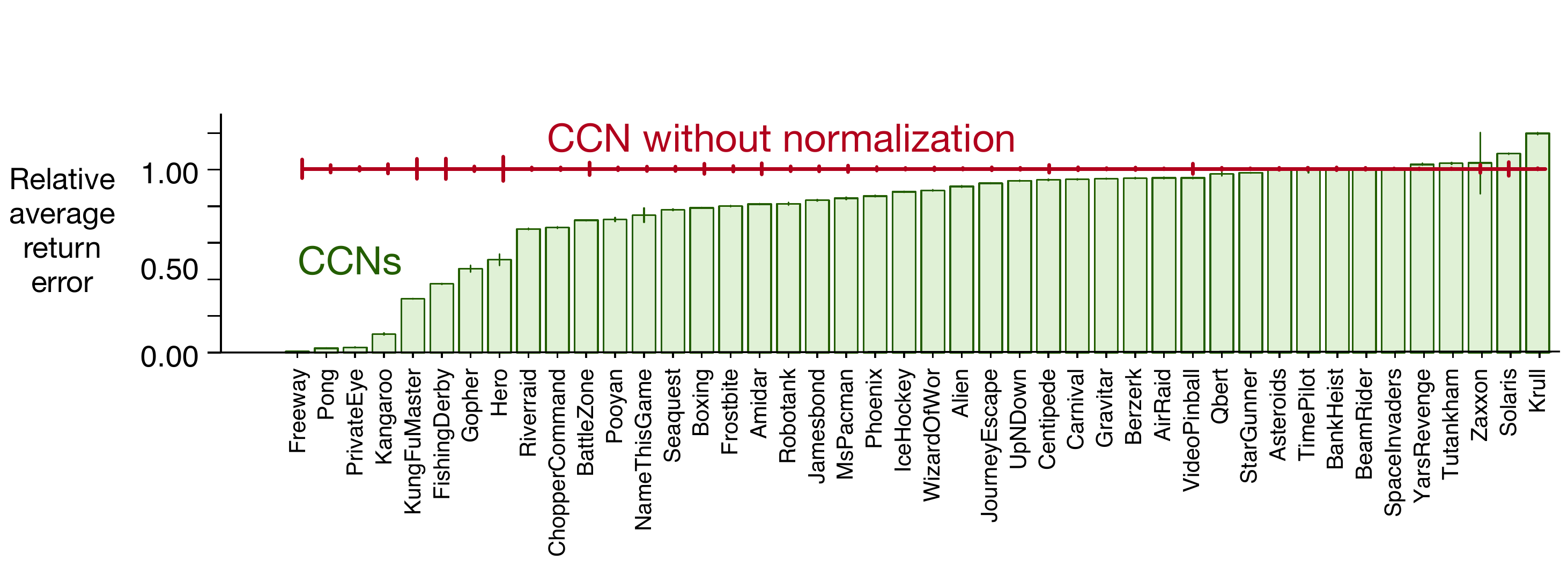}
	\caption{The impact of feature normalization on the performance of CCNs on the Atari prediction benchmark. We run both methods for 10 million steps and tune their hyper-parameters independently. The results exclude some games on which CCNs without normalization diverged. Feature normalization significantly improves the perfomance. All results are averaged over 15 runs and the error bars are +- standard error.}
	\label{atari_norm}
\end{figure*}

We do a similar experiment on the Atari prediction benchmark and report the performance of CCNs without normalization relative to CCNs in Figure~\ref{atari_norm} for ten million learning steps. We use CCNs with 4 features-per-stage, and 2.5 million steps-per-stage. We find that feature normalization improves the performance significantly on almost all games. 
\subsection{Impact of Hyperparameters on The Performance of CCNs}
\begin{figure}[t]
	\centering
	\includegraphics[width=0.99\textwidth]{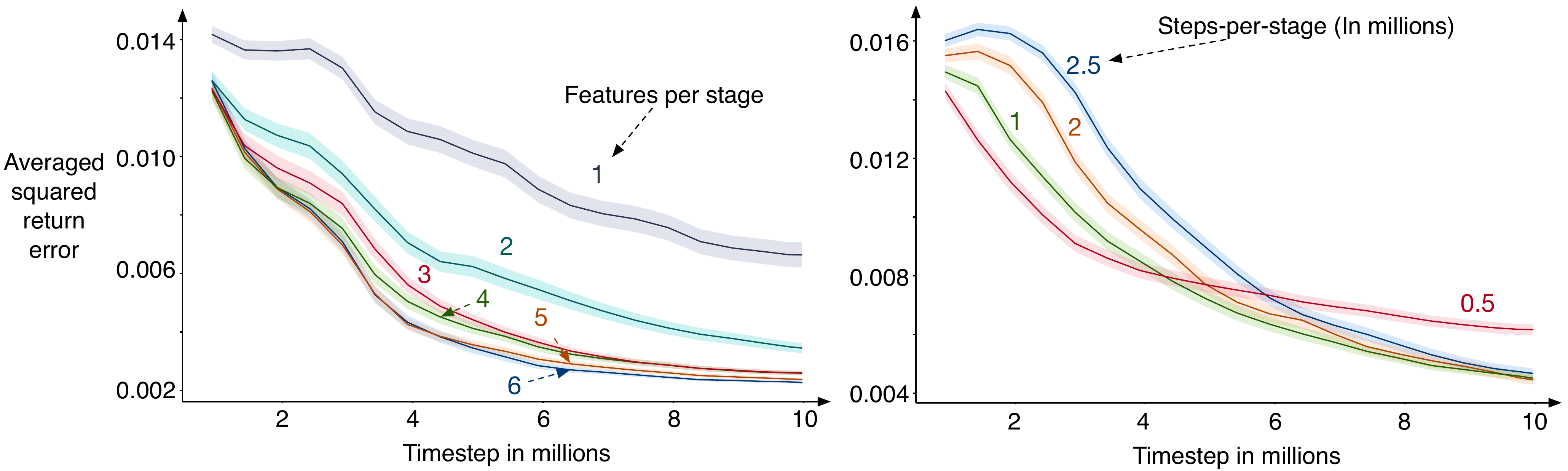}
	\caption{The impact of features-per-stage and steps-per-stage on the performance of CCNs.}
		\label{ccn_abl}
\end{figure}

We study the impact of the two important hyperparameters---\textit{steps-per-stage} and \textit{features-per-stage}---by conducting two experiments. In the first experiment, we fix the steps-per-stage to 2.5 million and vary the number of features from 1 to 6. We report the results in Figure~\ref{ccn_abl} (left). We use the same value of step-size---0.0001---for all curves. The results show that CCNs scale with more computation, and increasing the number of features-per-stage improves the performance monotonically. 

In the second experiment, we fix the features-per-stage to 2 and vary the steps-per-stage and report the results in Figure~\ref{ccn_abl} (right). The impact of steps-per-stage on performance is complex. A small steps-per-stage improves early learning, as the learner is adding more features quickly. However, if the value of steps-per-stage is too small---0.5 million in our experiment---the asymptotic performance is poor. Our guess is that freezing a feature too early is detrimental to the quality of the feature. An interesting future direction would be to automate when a feature is to be frozen. This could be done by waiting for a feature to converge before freezing it.

\subsection{Implementation Details}
We implement all methods in C++. For columnar, constructive, and CCN approaches, we use the update equations derived in Appendix~B. We verify the correctness of the gradients computed by our derived equations, and our implementation of T-BPTT by comparing them to the gradients computed by PyTorch for networks initialized to have the same parameters. The gradients given by our implementation and those by PyTorch match exactly. Our C++ implementation avoids the overhead of Python and PyTorch, and is around 50x faster for small recurrent networks as compared to PyTorch. Having a fast and efficient implementation was crucial for performing large hyperparameter sweeps and reporting statistically significant results by averaging over multiple seeds. 
\subsection{Compute Infrastructure used for Experiments}
We run all experiments on large CPU clusters. A single run of the trace patterning task for 10 million steps takes around 1 minutes on a single CPU, whereas a single run on Atari for 30 million steps takes around 2 hours. Both experiments take less than 2 GB of ram per run. We used 1,000 CPUs spread across the Cedar, Narval, and Beluga clusters provided by The Digital Research Alliance of Canada for running the experiments. 

\subsection{Equations for Estimating Compute Used by Each Method}
Every method uses roughly the same amount of computation per-step. We estimate the amount of compute used by each method by looking at its architecture and the learning algorithm. These estimates are not exact, and there may be some minor differences depending on how these methods are implemented in practice. However, the principle largely remains the same. And we have verified from our empirical observations that these estimates are close to what we observe.

 Let $|h|$ be the number of hidden features, $|x|$ be the number of input features, $k$ be the truncation length, and $u$ be the features-per-stage parameter. Then the total amount of computation used by an LSTM cell for a single forward pass can be estimated using the following equation: 
\begin{equation*}
    4|h| + 4|x| + 4
\end{equation*}
where the number four is due to the four gates used by an LSTM cell. In T-BPTT, we used a fully connected LSTM so the total number of features would be $|h|$. Forward pass of a fully connected LSTM would use:
\begin{equation*}
   |h|(4|h| + 4|x| + 4) = 4 |h|^2 + 4|h||x| + 4|h|
\end{equation*}
operations. Finally, T-BPTT requires k times more computation for computing the gradient, bringing the total cost to: 
\begin{align*}
   & 4 |h|^2 + 4|h||x| + 4|h| + k(4 |h|^2 + 4|h||x| + 4|h|) \\
   =& (k+1)(4 |h|^2 + 4|h||x| + 4|h|)
\end{align*}
For columnar, constructive and CCN, first we see from Appendix B that recursively computing the gradient is roughly six times more expensive than the forward pass of the LSTM, which, according to our empirical observations, is an overestimation. The number six comes from the observation that TC and TW are used in six operations for each parameter when estimating gradients. All computations other than those involving TC and TW can be cached and reused.
Total compute used by a single columnar cell for the forward pass, therefore, is: 
\begin{align*}
    4 + 4|x| + 4
\end{align*}
since hidden state = 1 for a single column. Compute used by $|h|$ cells is:
\begin{align*}
    |h|(4|x| + 8).
\end{align*}
Adding compute used by the learning algorithm, we get: 
\begin{align*}
    |h|(4|x| + 8) +  6|h|(4|x| + 8)
\end{align*}
In the CCN approach, on average, an LSTM cell takes as input $\frac{|h|}{2}$ hidden states. As a result, the compute used for a single forward pass by a single recurrent feature is given by: 
\begin{align*}
        4\frac{|h|}{2} + 4|x| + 4,
\end{align*}
and for $|h|$ features it is: 
\begin{align*}
        |h|(2|h| + 4|x| + 4).
\end{align*}
Since we learn $u$ features at a time, the total estimated compute per step for CCN networks is given by: 
\begin{align*}
    |h|(2|h| + 4|x| + 4) + 6u(2|h| + 4|x| + 4).
\end{align*}
For constructive networks, we can substitute $u=1$ in the equation above.
\section{Forward-mode Gradient Computation for an LSTM Cell }
\label{update_equations}
Here we derive the update equations for recursively computing the gradients of a single LSTM based recurrent column. Each column has a single hidden unit. Because all columns are identical, the same update equations can be used for learning in columnar, constructive, and the CCN approach. We compared the gradients estimated using the derived equations with the gradient computed using BPTT in PyTorch without truncation on random trajectories, and found them to match exactly. 

The state of an LSTM column is updated using following equations: 
\begin{align}
i(t) &= \sigma( W_{i}^T x_k(t) + u_{i} h(t-1) + b_i) \label{i} \\
f(t) &= \sigma( W_{f}^T x_k(t) + u_{f} h(t-1) + b_f) \label{f} \\
o(t) &= \sigma( W_{o}^T x_k(t) + u_{o} h(t-1) + b_o) \label{o} \\
g(t) &= \phi( W_{g}^T x_k(t) + u_{g} h(t-1) + b_g) \label{g} \\
c(t) &= f(t)  c(t-1) + i(t) g(t) \label{c} \\
h(t) &= o(t)  \phi(c(t)) \label{state_update}
\end{align}
where $\sigma$ and $\phi$ are the sigmoid and tanh activation functions, $h(t)$ is the state of the column at time $t$ and $ W_{i}^T x_k(t) = \sum_{k=1}^m W_{i_k} x_k(t)$. The derivative of $\sigma(x)$ and $\phi(x)$ w.r.t to $x$ are $\sigma(x)(1-\sigma(x))$ and $(1-\phi^2(x))$ respectively.

Let the length of input vector $x$ be $m$. Then, $W_{i}, W_{f}, W_{o}$ and $W_{g}$ are vectors of length $m$ whereas $u_i, b_i, u_f, b_f, u_o, b_o, u_g$ and $b_g$ are scalars. We want to compute gradient of $h(t)$ with respect to all the parameters. We derive the update equations for $\frac{\partial h(t)}{\partial W_{i}} ,\frac{\partial h(t)}{\partial u_{i}}, $$\frac{\partial h(t)}{\partial b_{i}}, \frac{\partial h(t)}{\partial W_{f}} $$,\frac{\partial h(t)}{\partial u_{f}}, \frac{\partial h(t)}{\partial b_{f}}, $$\frac{\partial h(t)}{\partial W_{o}} ,\frac{\partial h(t)}{\partial u_{o}},$$ \frac{\partial h(t)}{\partial b_{o}}, \frac{\partial h(t)}{\partial W_{g}} ,\frac{\partial h(t)}{\partial u_{g}},$ and $\frac{\partial h(t)}{\partial b_{g}}$ in the following sections. 

\subsection{$\frac{\partial h(t)}{\partial W_{i}}$}

$W_i = (W_{i_1}, W_{i_2}, \cdots, W_{i_m})$ is a vector of length $m$. Since all elements of $W_i$ are symmetric, we show gradient derivation for $W_{i_j}$ without loss of generality. Let

\begin{align}
TH_{W_{i_j}}(t)  & \vcentcolon = \frac{\partial h(t)}{\partial W_{i_j}}  && \textit{(By definition)} \label{def1} \\
TH_{W_{i_j}}(0)  & \vcentcolon = 0  && \textit{(By definition)} \label{def2} \\
TC_{W_{i_j}}(t)  & \vcentcolon = \frac{\partial c(t)}{\partial W_{i_j}}  && \textit{(By definition)} \label{def3} \\
TC_{W_{i_j}}(0)  & \vcentcolon = 0  && \textit{(By definition)}  \label{def4}
\end{align}
Then:
\begin{align*}
TH_{W_{i_j}}(t) & = \frac{\partial}{\partial W_{i_j}} \left ( o(t) \phi(c(t)) \right ) && \textit{From equation \ref{state_update} and definition \ref{def1}} \\
& = o(t) \frac{\partial \phi(c(t))}{\partial W_{i_j}}  + \phi(c(t)) \frac{\partial o(t)}{\partial W_{i_j}}   && \textit{Product rule of differentiation} \\
&= o(t) (1-\phi^{2}(c(t))) \frac{\partial c(t)}{\partial W_{i_j}}  + \phi(c(t)) \frac{\partial o(t)}{\partial W_{i_j}}   && \textit{Derivative of $\phi(x)$ is (1-$\phi^2(x))$}\\ 
&= o(t) (1-\phi^{2}(c(t))) TC_{W_{i_j}}(t)  + \phi(c(t)) \frac{\partial o(t)}{\partial W_{i_j}}   && \textit{From definition \ref{def3}}\\ 
\frac{\partial o(t)}{\partial W_{i_j}} &=  \frac{\partial}{\partial W_{i_j}} \sigma(W_{o}^T x(t) + u_{o} h(t-1) + b_o) && \textit{From equation \ref{o}}\\
&= \sigma(y)(1-\sigma(y))u_o TH_{W_{i_j}}(t-1) && \textit{Where $y$ equals $W_{o}^T x(t) + u_{o} h(t-1) + b_o$} \\
TC_{W_{i_j}}(t) &=  \frac{\partial c(t)}{\partial W_{i_j}} && \textit{From definition \ref{def3}} \\ 
&=  \frac{\partial}{\partial W_{i_j}} (f(t)  c(t-1) + i(t) g(t))  && \textit{From equation \ref{c}} \\
&= f(t) TC_{W_{i_j}}(t-1) + c(t-1)\frac{\partial f(t)}{ \partial W_{i_j}}    && \textit{Product rule and definition \ref{def3}}\\
&+ \frac{\partial}{\partial W_{i_j}} ( i(t) g(t) )\\
&= f(t) TC_{W_{i_j}}(t-1) + c(t-1)\frac{\partial f(t)}{ \partial W_{i_j}}   && \textit{Product rule} \\
&+ i(t)\frac{\partial g(t) }{\partial W_{i_j}}  + g(t)\frac{\partial i(t) }{\partial W_{i_j}}
\end{align*}

Where gradient of $g(t)$  w.r.t $W_{i_j}$ is:
\begin{align*}
\frac{\partial g(t)}{\partial W_{i_j}} &=  \frac{\partial}{\partial W_{i_j}} \phi(W_{g}^T x(t) + u_{g} h(t-1) + b_g) && \textit{From equation \ref{g}} \\
&= (1-\phi^2(y))u_g TH_{W_{i_j}}(t-1) && \textit{Where $y$ equals $W_{g}^T x(t) + u_{g} h(t-1) + b_g$} \\
\end{align*}
, gradient of $f(t)$  w.r.t $W_{i_j}$ is:
\begin{align*}
\frac{\partial f(t)}{\partial W_{i_j}} &=  \frac{\partial}{\partial W_{i_j}} \sigma(W_{f}^T x(t) + u_{f} h(t-1) + b_f) && \textit{From equation \ref{f}} \\
&= \sigma(y)(1-\sigma(y))u_f TH_{W_{i_j}}(t-1) && \textit{Where $y$ equals $W_{f}^T x(t) + u_{f} h(t-1) + b_f$} \\
\end{align*}
and gradient of $i(t)$ w.r.t $W_{i_j}$ is:
\begin{align*}
\frac{\partial i(t)}{\partial W_{i_j}} &=  \frac{\partial}{\partial W_{i_j}} \sigma(W_{i}^T x(t) + u_{i} h(t-1) + i_f) && \textit{From equation \ref{i}} \\
&= \sigma(y)(1-\sigma(y)) \left (x_j(t)+ u_i TH_{W_{i_j}}(t-1) \right )&& \textit{Where $y$ equals $W_{i}^T x(t) + u_{i} h(t-1) + b_i$} \\
\end{align*}
The derivation shows that using two traces per parameter of $W_i$, it is possible to compute the gradient of $h(t)$ w.r.t $W_i$ recursively. We provide the derivations for parameters $u_i$ and $b_i$ below. We skip the step-by-step derivations for the remaining parameters as they are similar. 

\subsection{$\frac{\partial h(t)}{\partial u_i}$}

\begin{align}
TH_{u_{i}}(t)  & \vcentcolon = \frac{\partial h(t)}{\partial u_i}  && \textit{(By definition)}  \label{thu}\\
TH_{u_{i}}(0)  & \vcentcolon = 0  && \textit{(By definition)}  \\
TC_{u_{i}}(t)  & \vcentcolon = \frac{\partial c(t)}{\partial u_i}  && \textit{(By definition)}  \label{tcu} \\
TC_{u_{i}}(0)  & \vcentcolon = 0  && \textit{(By definition)} 
\end{align}

\begin{align*}
TH_{u_i}(t) & = \frac{\partial}{\partial u_i} \left ( o(t) \phi(c(t)) \right ) && \textit{From equation \ref{state_update}} \\
& = o(t) \frac{\partial \phi(c(t))}{\partial u_i}  + \phi(c(t)) \frac{\partial o(t)}{\partial u_i}   && \textit{Product rule} \\
&= o(t) (1-\phi^{2}(c(t))) \frac{\partial c(t)}{\partial u_i}  + \phi(c(t)) \frac{\partial o(t)}{\partial u_i}   && \textit{Derivative of $\phi(x)$ is $1-\phi^2(x)$} \\
&= o(t) (1-\phi^{2}(c(t))) TC_{u_{i}}(t)  + \phi(c(t)) \frac{\partial o(t)}{\partial u_i}   && \textit{Using definition \ref{tcu}} \\
\frac{\partial o(t)}{\partial u_i} &=  \frac{\partial}{\partial u_i} \sigma(W_{o}^T x(t) + u_{o} h(t-1) + b_o) && \textit{Using equations \ref{o}}\\
&= \sigma(x)(1-\sigma(x))u_o TH_{u_i}(t-1) && \textit{Where $x$ equal $W_{o}^T x(t) + u_{o} h(t-1) + b_o$} \\
TC_{u_i}(t) &=  \frac{\partial c(t)}{\partial u_i} && \textit{Definition \ref{tcu}} \\ 
&=  \frac{\partial}{\partial u_i} (f(t)  c(t-1) + i(t) g(t))  && \textit{From equation \ref{state_update}} \\
&= f(t) TC_{u_i}(t-1) + c(t-1)\frac{\partial f(t)}{ \partial u_i}  && \textit{Product rule}\\
&+ \frac{\partial}{\partial u_i} \left ( i(t) g(t) ) \right ) \\
&= f(t) TC_{u_i}(t-1) + c(t-1)\frac{\partial f(t)}{ \partial u_i}  && \textit{Product rule} \\
&+ i(t)\frac{\partial g(t) }{\partial u_i}  + g(t)\frac{\partial i(t) }{\partial u_i}
\end{align*}
Gradient of  $g(t)$  w.r.t  $u_i$ is:
\begin{align*}
\frac{\partial g(t)}{\partial u_i} &=  \frac{\partial}{\partial u_i} \phi(W_{g}^T x(t) + u_{g} h(t-1) + b_g) && \textit{From equations \ref{state_update}} \\
&= (1-\phi^2(y))u_g TH_{u_i}(t-1) && \textit{Where $y$ equals $W_{g}^T x(t) + u_{g} h(t-1) + b_g$} \\
\end{align*}
, gradient of $f(t)$  w.r.t $u_i$ is:
\begin{align*}
\frac{\partial f(t)}{\partial u_i} &=  \frac{\partial}{\partial u_i} \sigma(W_{f}^T x(t) + u_{f} h(t-1) + b_f) && \textit{From equations 1} \\
&= \sigma(y)(1-\sigma(y))u_f TH_{u_i}(t-1) && \textit{Where $y$ equals $W_{f}^T x(t) + u_{f} h(t-1) + b_f$} \\
\end{align*}
and the gradient of $i(t)$  w.r.t $u_i$ is

\begin{align*}
\frac{\partial i(t)}{\partial u_i} &=  \frac{\partial}{\partial u_i} \sigma(W_{i}^T x(t) + u_{i} h(t-1) + b_i) && \textit{Using equations 1} \\
&= \sigma(y)(1-\sigma(y)) \left (h(t-1)+ u_i TH_{u_{i}}(t-1) \right )&& \textit{Where $y$ equals $W_{i}^T x(t) + u_{i} h(t-1) + b_i$} \\
\end{align*}
% \end{equation}

\subsection{$\frac{\partial h(t)}{\partial b_i}$}

% \begin{equation}
\begin{align}
TH_{b_i}(t)  & \vcentcolon = \frac{\partial h(t)}{\partial b_i}  && \textit{(By definition)}  \label{thb}\\
TH_{b_i}(0)  & \vcentcolon = 0  && \textit{(By definition)}  \\
TC_{b_i}(t)  & \vcentcolon = \frac{\partial c(t)}{\partial b_i}  && \textit{(By definition)}  \label{tcb}\\
TC_{b_i}(0)  & \vcentcolon = 0  && \textit{(By definition)} 
\end{align}
% \end{equation}

% \begin{equation}
\begin{align*}
TH_{b_i}(t) & = \frac{\partial}{\partial b_i} \left ( o(t) \phi(c(t)) \right ) && \textit{From equation \ref{state_update}} \\
& = o(t) \frac{\partial \phi(c(t))}{\partial b_i}  + \phi(c(t)) \frac{\partial o(t)}{\partial b_i}   && \textit{Product rule} \\
&= o(t) (1-\phi^{2}(c(t))) \frac{\partial c(t)}{\partial b_i}  + \phi(c(t)) \frac{\partial o(t)}{\partial b_i}   && \textit{Derivative of of $\phi(x)$ is $1-\phi^2(x)$}\\ 
&= o(t) (1-\phi^{2}(c(t))) TC_{b_i}(t)  + \phi(c(t)) \frac{\partial o(t)}{\partial b_i}   && \textit{From definition \ref{tcb} }\\ 
\frac{\partial o(t)}{\partial b_i} &=  \frac{\partial}{\partial b_i} \sigma(W_{o}^T x(t) + u_{o} h(t-1) + b_o) && \textit{From equations \ref{o}}\\
&= \sigma(y)(1-\sigma(y))u_o TH_{b_i}(t-1) && \textit{Where $y$ equal $W_{o}^T x(t) + u_{o} h(t-1) + b_o$} \\
TC_{b_i}(t) &=  \frac{\partial c(t)}{\partial b_i} && \textit{From definition \ref{tcb}} \\ 
&=  \frac{\partial}{\partial b_i} (f(t)  c(t-1) + i(t) g(t))  && \textit{From equation \ref{c}} \\
&= f(t) TC_{b_i}(t-1) + c(t-1)\frac{\partial f(t)}{ \partial b_i}   && \textit{Product rule}\\
&+ \frac{\partial}{\partial b_i}  i(t) g(t) \\
&= f(t) TC_{b_i}(t-1) + c(t-1)\frac{\partial f(t)}{ \partial b_i} && \textit{Product rule} \\
& + i(t)\frac{\partial g(t) }{\partial b_i}  + g(t)\frac{\partial i(t) }{\partial b_i} 
\end{align*}
% \end{equation}
Where gadient of  $g(t)$  w.r.t $b_i$ is:
\begin{align*}
\frac{\partial g(t)}{\partial b_i} &=  \frac{\partial}{\partial b_i} \phi(W_{g}^T x(t) + u_{g} h(t-1) + b_g) && \textit{From equation \ref{g}} \\
&= (1-\phi^2(y))u_g TH_{b_i}(t-1) && \textit{Where $y$ equal $W_{g}^T x(t) + u_{g} h(t-1) + b_g$} \\
\end{align*}

, gradient of $f(t)$  w.r.t $b_i$ is:
% \begin{equation}
\begin{align*}
\frac{\partial f(t)}{\partial b_i} &=  \frac{\partial}{\partial b_i} \sigma(W_{f}^T x(t) + u_{f} h(t-1) + b_f) && \textit{From equation \ref{f}} \\
&= \sigma(y)(1-\sigma(y))u_f TH_{b_i}(t-1) && \textit{Where $y$ equal $W_{f}^T x(t) + u_{f} h(t-1) + b_f$} \\
\end{align*}
% \end{equation}

and gradient of $i(t)$  w.r.t $b_i$ is:
% \begin{equation}
\begin{align*}
\frac{\partial i(t)}{\partial b_i} &=  \frac{\partial}{\partial b_i} \sigma(W_{i}^T x(t) + u_i h(t-1) + b_i) && \textit{From equation \ref{i}} \\
&= \sigma(y)(1-\sigma(y)) \left (u_i TH_{b_i}(t-1) + 1 \right )&& \textit{Where $y$ equal $W_{i}^T x(t) + b_i h(t-1) + b_i$} \\
\end{align*}
% \end{equation}

\subsection{$\frac{\partial h(t)}{\partial W_{f_j}}$}
The derivations for the remaining parameters is analogous to what previous derivations. The final equations are as follows.
\begin{equation}
\begin{aligned}
\frac{\partial g(t)}{\partial W_{f_j}} &=  (1-\phi^2(y))(u_g TH_{W_{f_j}}(t-1))  \\
\frac{\partial f(t)}{\partial W_{f_j}} &=  \sigma(y)(1-\sigma(y))( x_j + u_f TH_{W_{f_j}}(t-1)) \\
\frac{\partial i(t)}{\partial W_{f_j}}  &= \sigma(y)(1-\sigma(y))( u_i TH_{W_{f_j}}(t-1))  \\
\frac{\partial o(t)}{\partial W_{f_j}} &=  \sigma(y)(1-\sigma(y))( u_o TH_{W_{f_j}}(t-1))  \\
TC_{W_{f_j}}&= f(t) TC_{f_j}(t-1) + c(t-1)\frac{\partial f(t)}{ \partial b_i} + i(t)\frac{\partial g(t) }{\partial b_i}  + g(t)\frac{\partial i(t) }{\partial b_i}  \\
TH_{W_{f_j}}&= o(t) (1-\phi^{2}(c(t))) TC_{W_{f_j}}(t)  + \phi(c(t)) \frac{\partial o(t)}{\partial W_{ij}}  
\end{aligned}
\end{equation}

\subsection{$\frac{\partial h(t)}{\partial W_{o_j}}$}

\begin{equation}
\begin{aligned}
\frac{\partial g(t)}{\partial W_{o_j}} &=  (1-\phi^2(y))(u_g TH_{W_{o_j}}(t-1))  \\
\frac{\partial f(t)}{\partial W_{o_j}} &=  \sigma(y)(1-\sigma(y))( u_f TH_{W_{o_j}}(t-1)) \\
\frac{\partial i(t)}{\partial W_{o_j}}  &= \sigma(y)(1-\sigma(y)) u_i TH_{W_{o_j}}(t-1)  \\
\frac{\partial o(t)}{\partial W_{o_j}} &=  \sigma(x)(1-\sigma(x))(x_j + u_o TH_{W_{o_j}}(t-1))  \\
TC_{W_{o_j}}&= f(t) TC_{o_j}(t-1) + c(t-1)\frac{\partial f(t)}{ \partial b_i} + i(t)\frac{\partial g(t) }{\partial b_i}  + g(t)\frac{\partial i(t) }{\partial b_i}  \\
TH_{W_{o_j}}&= o(t) (1-\phi^{2}(c(t))) TC_{W_{o_j}}(t)  + \phi(c(t)) \frac{\partial o(t)}{\partial W_{ij}}  
\end{aligned}
\end{equation}

\subsection{$\frac{\partial h(t)}{\partial W_{g_j}}$}

\begin{equation}
\begin{aligned}
\frac{\partial g(t)}{\partial W_{g_j}} &=  (1-\phi^2(y))(x_j + u_g TH_{W_{g_j}}(t-1))  \\
\frac{\partial f(t)}{\partial W_{g_j}} &=  \sigma(y)(1-\sigma(y))( u_f TH_{W_{g_j}}(t-1)) \\
\frac{\partial i(t)}{\partial W_{g_j}}  &= \sigma(y)(1-\sigma(y))(u_i TH_{W_{g_j}}(t-1))  \\
\frac{\partial o(t)}{\partial W_{g_j}} &=  \sigma(x)(1-\sigma(x))( u_o TH_{W_{g_j}}(t-1))  \\
TC_{W_{g_j}}&= f(t) TC_{g_j}(t-1) + c(t-1)\frac{\partial f(t)}{ \partial b_i} + i(t)\frac{\partial g(t) }{\partial b_i}  + g(t)\frac{\partial i(t) }{\partial b_i}  \\
TH_{W_{g_j}}&= o(t) (1-\phi^{2}(c(t))) TC_{W_{g_j}}(t)  + \phi(c(t)) \frac{\partial o(t)}{\partial W_{ij}}  
\end{aligned}
\end{equation}

\subsection{$\frac{\partial h(t)}{\partial u_{o}}$}

\begin{equation}
\begin{aligned}
\frac{\partial g(t)}{\partial u_{o}} &=  (1-\phi^2(y))(u_g TH_{u_{o}}(t-1))  \\
\frac{\partial f(t)}{\partial u_{o}} &=  \sigma(y)(1-\sigma(y))( u_f TH_{u_{o}}(t-1)) \\
\frac{\partial i(t)}{\partial u_{o}}  &= \sigma(y)(1-\sigma(y))(u_i TH_{u_{o}}(t-1))  \\
\frac{\partial o(t)}{\partial u_{o}} &=  \sigma(x)(1-\sigma(x))( u_o TH_{u_{o}}(t-1) + h(t-1))  \\
TC_{u_{o}}&= f(t) TC_{i_j}(t-1) + c(t-1)\frac{\partial f(t)}{ \partial b_i} + i(t)\frac{\partial g(t) }{\partial b_i}  + g(t)\frac{\partial i(t) }{\partial b_i}  \\
TH_{u_{o}}&= o(t) (1-\phi^{2}(c(t))) TC_{u_{o}}(t)  + \phi(c(t)) \frac{\partial o(t)}{\partial W_{ij}}  
\end{aligned}
\end{equation}

\subsection{$\frac{\partial h(t)}{\partial u_{f}}$}

\begin{equation}
\begin{aligned}
\frac{\partial g(t)}{\partial u_{f}} &=  
(1-\phi^2(y))(u_g TH_{u_{f}}(t-1))  \\
\frac{\partial f(t)}{\partial u_{f}} &=  \sigma(y)(1-\sigma(y))( u_f TH_{u_{f}}(t-1) + h(t-1)) \\
\frac{\partial i(t)}{\partial u_{f}}  &= \sigma(y)(1-\sigma(y))(u_i TH_{u_{f}}(t-1))  \\
\frac{\partial o(t)}{\partial u_{f}} &=  \sigma(x)(1-\sigma(x))( u_o TH_{u_{f}}(t-1))  \\
TC_{u_{f}}&= f(t) TC_{i_j}(t-1) + c(t-1)\frac{\partial f(t)}{ \partial b_i} + i(t)\frac{\partial g(t) }{\partial b_i}  + g(t)\frac{\partial i(t) }{\partial b_i}  \\
TH_{u_{f}}&= o(t) (1-\phi^{2}(c(t))) TC_{u_{f}}(t)  + \phi(c(t)) \frac{\partial o(t)}{\partial W_{ij}}  
\end{aligned}
\end{equation}

\subsection{$\frac{\partial h(t)}{\partial u_{g}}$}

\begin{equation}
\begin{aligned}
\frac{\partial g(t)}{\partial u_{g}} &=  
(1-\phi^2(y))(u_g TH_{u_{g}}(t-1)+ h(t-1))  \\
\frac{\partial f(t)}{\partial u_{g}} &=  \sigma(y)(1-\sigma(y))( u_f TH_{u_{g}}(t-1)) \\
\frac{\partial i(t)}{\partial u_{g}}  &= \sigma(y)(1-\sigma(y))( u_i TH_{u_{g}}(t-1))  \\
\frac{\partial o(t)}{\partial u_{g}} &=  \sigma(x)(1-\sigma(x))( u_o TH_{u_{g}}(t-1))  \\
TC_{u_{g}}&= f(t) TC_{i_j}(t-1) + c(t-1)\frac{\partial f(t)}{ \partial b_i} + i(t)\frac{\partial g(t) }{\partial b_i}  + g(t)\frac{\partial i(t) }{\partial b_i}  \\
TH_{u_{g}}&= o(t) (1-\phi^{2}(c(t))) TC_{u_{g}}(t)  + \phi(c(t)) \frac{\partial o(t)}{\partial W_{ij}}  
\end{aligned}
\end{equation}

\subsection{$\frac{\partial h(t)}{\partial b_{g}}$}

\begin{equation}
\begin{aligned}
\frac{\partial g(t)}{\partial b_{g}} &=  
(1-\phi^2(y))(u_g TH_{b_{g}}(t-1)+ 1)  \\
\frac{\partial f(t)}{\partial b_{g}} &=  \sigma(y)(1-\sigma(y))( u_f TH_{b_{g}}(t-1)) \\
\frac{\partial i(t)}{\partial b_{g}}  &= \sigma(y)(1-\sigma(y))(u_i TH_{b_{g}}(t-1))  \\
\frac{\partial o(t)}{\partial b_{g}} &=  \sigma(x)(1-\sigma(x))( u_o TH_{b_{g}}(t-1))  \\
TC_{b_{g}}&= f(t) TC_{i_j}(t-1) + c(t-1)\frac{\partial f(t)}{ \partial b_i} + i(t)\frac{\partial g(t) }{\partial b_i}  + g(t)\frac{\partial i(t) }{\partial b_i}  \\
TH_{b_{g}}&= o(t) (1-\phi^{2}(c(t))) TC_{b_{g}}(t)  + \phi(c(t)) \frac{\partial o(t)}{\partial W_{ij}}  
\end{aligned}
\end{equation}

\subsection{$\frac{\partial h(t)}{\partial b_{f}}$}

\begin{equation}
\begin{aligned}
\frac{\partial g(t)}{\partial b_{f}} &=  
(1-\phi^2(y))(u_g TH_{b_{f}}(t-1))  \\
\frac{\partial f(t)}{\partial b_{f}} &=  \sigma(y)(1-\sigma(y))( u_f TH_{b_{f}}(t-1) + 1 ) \\
\frac{\partial i(t)}{\partial b_{f}}  &= \sigma(y)(1-\sigma(y))( u_i TH_{b_{f}}(t-1))  \\
\frac{\partial o(t)}{\partial b_{f}} &=  \sigma(x)(1-\sigma(x))( u_o TH_{b_{f}}(t-1))  \\
TC_{b_{f}}&= f(t) TC_{i_j}(t-1) + c(t-1)\frac{\partial f(t)}{ \partial b_i} + i(t)\frac{\partial g(t) }{\partial b_i}  + g(t)\frac{\partial i(t) }{\partial b_i}  \\
TH_{b_{f}}&= o(t) (1-\phi^{2}(c(t))) TC_{b_{f}}(t)  + \phi(c(t)) \frac{\partial o(t)}{\partial W_{ij}}  
\end{aligned}
\end{equation}
\subsection{$\frac{\partial h(t)}{\partial b_{o}}$}
\begin{equation}
\begin{aligned}
\frac{\partial g(t)}{\partial b_{o}} &=  
(1-\phi^2(y))(u_g TH_{b_{o}}(t-1))  \\
\frac{\partial f(t)}{\partial b_{o}} &=  \sigma(y)(1-\sigma(y))( u_f TH_{b_{o}}(t-1)) \\
\frac{\partial i(t)}{\partial b_{o}}  &= \sigma(y)(1-\sigma(y))( u_i TH_{b_{o}}(t-1))  \\
\frac{\partial o(t)}{\partial b_{o}} &=  \sigma(x)(1-\sigma(x))( u_o TH_{b_{o}}(t-1) + 1 )  \\
TC_{b_{o}}&= f(t) TC_{i_j}(t-1) + c(t-1)\frac{\partial f(t)}{ \partial b_i} + i(t)\frac{\partial g(t) }{\partial b_i}  + g(t)\frac{\partial i(t) }{\partial b_i}  \\
TH_{b_{o}}&= o(t) (1-\phi^{2}(c(t))) TC_{b_{o}}(t)  + \phi(c(t)) \frac{\partial o(t)}{\partial W_{ij}}  
\end{aligned}
\end{equation}

\vskip 0.2in

\section*{References}

\hangin Bellemare, M. G., Naddaf, Y., Veness, J., \& Bowling, M. (2013). The arcade learning environment: An evaluation platform for general agents. The journal of artificial intelligence research.
\hangin Cooijmans, T., Ballas, N., Laurent, C., Gülçehre, Ç., \& Courville, A. (2017). Recurrent batch normalization.  International conference on learning representations.
\hangin Dohare, S., Hernandez-Garcia, J. F., Rahman, P., Sutton, R. S., \& Mahmood, A. R. (2023). Maintaining Plasticity in Deep Continual Learning. arXiv preprint arXiv:2306.13812.
\hangin Elman, J. L. (1990). Finding structure in time. Cognitive science.
\hangin Fahlman, S. (1990). The recurrent cascade-correlation architecture. Advances in neural information processing systems.
\hangin Fujita, Y., Nagarajan, P., Kataoka, T., \& Ishikawa, T. (2021). Chainerrl: A deep reinforcement learning library. The journal of machine learning research.
\hangin Giles, C. L., Chen, D., Sun, G. Z., Chen, H. H., Lee, Y. C., \& Goudreau, M. W. (1995). Constructive learning of recurrent neural networks: limitations of recurrent cascade correlation and a simple solution. IEEE Transactions on Neural Networks.
\hangin Hessel, M., Modayil, J., Van Hasselt, H., Schaul, T., Ostrovski, G., Dabney, W., ... \& Silver, D. (2018). Rainbow: Combining improvements in deep reinforcement learning. AAAI conference on artificial intelligence.
\hangin Hochreiter, S., \& Schmidhuber, J. (1997). Long short-term memory. Neural computation.
\hangin Ioffe, S., \& Szegedy, C. (2015). Batch normalization: Accelerating deep network training by reducing internal covariate shift. International conference on machine learning.
\hangin Kapturowski, S., Ostrovski, G., Quan, J., Munos, R., \& Dabney, W. (2019). Recurrent experience replay in distributed reinforcement learning. International conference on learning representations.
\hangin Kremer, S. (1995). Finite state automata that recurrent cascade-correlation cannot represent. Advances in neural information processing systems.
\hangin Machado, M. C., Bellemare, M. G., Talvitie, E., Veness, J., Hausknecht, M., \& Bowling, M. (2018). Revisiting the arcade learning environment: Evaluation protocols and open problems for general agents. Journal of Artificial Intelligence Research.
\hangin Massé, P. Y., \& Ollivier, Y. (2020). Convergence of online adaptive and recurrent optimization algorithms. arXiv preprint arXiv:2005.05645.
\hangin Menick, J., Elsen, E., Evci, U., Osindero, S., Simonyan, K., \& Graves, A. (2021). A practical sparse approximation for real time recurrent learning.  International conference on learning representations.
\hangin Mikolov, T., Karafiát, M., Burget, L., Cernocký, J., \& Khudanpur, S. (2010). Recurrent neural network based language model. Interspeech.
\hangin Mikolov, T., Kopecky, J., Burget, L., \& Glembek, O. (2009). Neural network based language models for highly inflective languages. International conference on acoustics, speech and signal processing.
\hangin Mnih, V., Kavukcuoglu, K., Silver, D., Rusu, A. A., Veness, J., Bellemare, M. G., ... \& Hassabis, D. (2015). Human-level control through deep reinforcement learning. nature.
\hangin Mountcastle, V. B. (1957). Modality and topographic properties of single neurons of cat's somatic sensory cortex. Journal of neurophysiology.
\hangin Mujika, A., Meier, F., \& Steger, A. (2018). Approximating real-time recurrent learning with random kronecker factors. Advances in neural information processing systems.
\hangin Rafiee, B., Abbas, Z., Ghiassian, S., Kumaraswamy, R., Sutton, R., Ludvig, E., \& White, A. (2022). From Eye-blinks to State Construction: Diagnostic Benchmarks for Online Representation Learning. Adaptive behavior.
\hangin Robinson, A. J., \& Fallside, F. (1987). The utility driven dynamic error propagation network. University of Cambridge.
\hangin Rumelhart, D. E., Hinton, G. E., \& Williams, R. J. (1986). Learning representations by back-propagating errors. Nature.
\hangin Sutskever, I. (2013). Training recurrent neural networks. University of Toronto.
\hangin Sutton, R. S., \& Barto, A. G. (2018). Reinforcement learning: An introduction. MIT press.
\hangin Sutton, R. S. (1984). Temporal credit assignment in reinforcement learning. University of Massachusetts Amherst.
\hangin Sutton, R. S. (1988). Learning to predict by the methods of temporal differences. Machine learning.
\hangin Tallec, C., \& Ollivier, Y. (2018). Unbiased online recurrent optimization. International conference on learning representations.
\hangin Tange, O. (2018). GNU parallel 2018. Lulu. com.
\hangin Tesauro, G. (1995). Temporal difference learning and TD-Gammon. Communications of the ACM.
\hangin Vinyals, O., Babuschkin, I., Czarnecki, W. M., Mathieu, M., Dudzik, A., Chung, J., ... \& Silver, D. (2019). Grandmaster level in StarCraft II using multi-agent reinforcement learning. Nature.
\hangin Werbos, P. J. (1988). Generalization of backpropagation with application to a recurrent gas market model. Neural networks.
\hangin Williams, R. J., \& Zipser, D. (1989). A learning algorithm for continually running fully recurrent neural networks. Neural computation.
\hangin Williams, R. J., \& Peng, J. (1990). An efficient gradient-based algorithm for on-line training of recurrent network trajectories. Neural computation.

\end{document}